\documentclass[5p,twocolumn]{elsarticle}

\usepackage{graphicx}
 \graphicspath{{images/}}
\DeclareGraphicsExtensions{.pdf,.jpeg,.png,.eps}
\usepackage{epstopdf}

\usepackage{graphicx}
\usepackage{amsfonts}
\usepackage{amsmath}

\usepackage{array}
\usepackage{multirow}
\usepackage{tabularx,booktabs}
\usepackage{hyperref}

\usepackage{xcolor}
\usepackage{soul}

\newcolumntype{Y}{>{\centering\arraybackslash}X}
\newcolumntype{s}{>{\hsize=.69\hsize}X}
\newcolumntype{b}{>{\hsize=.3\hsize}X}
\newcolumntype{d}{>{\hsize=.55\hsize}X}

\begin{document}

\begin{frontmatter}

\title{Joint Segmentation and Classification of Retinal Arteries/Veins from Fundus Images\tnoteref{mytitlenote}}

\tnotetext[mytitlenote]{$\copyright$ 2019. This manuscript version is made available under the CC-BY-NC-ND 4.0 license http://creativecommons.org/licenses/by-nc-nd/4.0/}

\author[mymainaddress]{Fantin~Girard}
\author[mysecondaryaddress]{Conrad~Kavalec}
\author[mymainaddress]{Farida~Cheriet}

\address[mymainaddress]{ Polytechnique Montreal, Montreal, QC H3T 1J4, Canada}
\address[mysecondaryaddress]{St Mary’s Hospital, Montreal, QC H3T 1M5, Canada}


\begin{abstract}
\textit{ Objective:} Automatic artery/vein (A/V) segmentation from fundus images is required to track blood vessel changes occurring with many pathologies including retinopathy and cardiovascular pathologies. One of the clinical measures that quantifies vessel changes is the arterio-venous ratio (AVR) which represents the ratio between artery and vein diameters. This measure significantly depends on the accuracy of vessel segmentation and classification into arteries and veins. This paper proposes a fast, novel method for semantic A/V segmentation combining deep learning and graph propagation.\textit{ Methods:} A convolutional neural network (CNN) is proposed to jointly segment and classify vessels into arteries and veins. The initial CNN labeling is propagated through a graph representation of the retinal vasculature, whose nodes are defined as the vessel branches and edges are weighted by the cost of linking pairs of branches. To efficiently propagate the labels, the graph is simplified into its minimum spanning tree. \textit{ Results:} The method achieves an accuracy of 94.8\% for vessels segmentation. The A/V classification achieves a specificity of 92.9\% with a sensitivity of 93.7\% on the CT-DRIVE database compared to the state-of-the-art-specificity and sensitivity, both of 91.7\%.\textit{ Conclusion:} The results show that our method outperforms the leading previous works on a public dataset for A/V classification and is by far the fastest. \textit{ Significance:} The proposed global AVR calculated on the whole fundus image using our automatic A/V segmentation method can better track vessel changes associated to diabetic retinopathy than the standard local AVR calculated only around the optic disc.
\end{abstract}


\end{frontmatter}


\section{Introduction}

Blood supply to the retina nerve fibers is ensured by the small arteries and veins that compose the retinal vasculature. Changes in blood vessels occur with many systemic or environmental factors such as diabetic conditions, aging, cardiovascular abnormalities and smoking. These factors can lead to abnormal changes in the vessel diameters, which appear earliest on the smallest vessels. Non-invasive access to small vessels is possible by imaging the retinal arterioles and venules using fundus digital cameras.
To quantify the severity of the diameters changes, the arterio-venous ratio (AVR), defined as the ratio between the arteriolar and venular diameters and usually calculated in a region around the optic disc (OD) \cite{knudtson}, is used. The correlations between vessel measures (diameters and AVR) and systemic factors have been quantified in many studies \cite{suncong} like the Rotterdam and Wisconsin studies \cite{rotterdam, avrdr}. In these studies, a lower AVR was related to many factors such as intima-media thickness (IMT) and carotid plaque score (indicating risk of coronary artery disease), hypertension, cholesterol level, progression of retinopathy and smoking. Arteriolar narrowing is related to higher blood pressure and aging. A lower AVR does not depend only on narrowing artery diameters. In fact, for diabetic retinopathy, the lower AVR is mainly explained by venular widening. Larger venular diameters are indeed associated to atherosclerosis, aortic calcification, progression of diabetic retinopathy (DR), incidence of proliferative retinopathy and cholesterol level \cite{rotterdam}. Relevant vessel changes are difficult to be manually estimated in clinical practice as the procedure is laborious and time-consuming\cite{niemeijer}.

To analyze vessel changes, automatic measurement of the vessel diameters is therefore required and involves implementing two tasks: \textbf{a)} vessels segmentation for diameter measurement; and \textbf{b)} classification into arteries and veins in order to track changes specifically in veins or arteries and calculate the AVR.

Vessels segmentation in fundus images has been subject of a great deal of research. Extracting the vasculature is a useful tool for automatic analysis of the fundus images, for example to separate hemorrhages from the real vasculature or as mentioned before to measure changes in the vessels diameters \cite{abramoff}. Three classes of methods can be identified from the vessel segmentation literature reviews \cite{abramoff, fraz1}:  edge-based methods, region-based methods and classification-based methods.
Edge-based methods comprise matched filter techniques \cite{chaudhuri,zhang2}, Hessian filtering \cite{frangi, martina, budai} and line-detector techniques \cite{ricci, nguyen,mendonca}. Matched filtering consists in convolving the image with a multiscale kernel that models the structure to be segmented. Vessels may be simply modeled by a Gaussian \cite{chaudhuri} or more complex distribution that accounts for light reflections or bifurcations \cite{wang,girardomia}. The subsequent thresholding can be adaptive, like hysteresis thresholding \cite{zhang2, roychowdhury, budai}, and be postprocessed with morphological operations \cite{miri}. Derivative filtering like Hessian filtering uses the tubular aspect of blood vessels \cite{frangi, martina} and can be combined with matched filter techniques \cite{sofka}.
Line-detector filters model the vessels as multiscale lines \cite{lazar}; bifurcations can then be reconstructed by analyzing the polar response of the filter \cite{you}.
As a second step to reduce the false negatives, region-growing techniques are often used by aggregating pixels of similar intensities along the vessel orientations \cite{roychowdhury, mendonca, nayebifar}. Classification-based segmentation methods are used either directly or to reduce false positives after applying edge-based techniques. The classification of pixels into vessels and background can be either unsupervised, with pixels being clustered into the two classes using K-means or self-organizing maps \cite{lupascu}, or supervised. In the latter case, labeled data on a training dataset are needed. The features used for training are often derived from edge-based techniques: intensity and derivative features \cite{staal, marin, niemeijer, fraz2}, or line-detector features \cite{ricci}. Many classifiers have been used, such as K-Nearest Neighbors (K-NN) \cite{niemeijer, staal}, support vector machines (SVM) \cite{ricci}, decision trees \cite{fraz2} and neural networks \cite{marin}.

Most of artery/vein classification methods (A/V) learn local features with machine learning techniques. Then, the initial machine learning classification is usually improved by exploiting the structural information from the vascular tree. \cite{niemeijer,epenhoff, dasht}. Graph-based analysis that introduces global topology rules is a common approach to correctly trace branches in the vascular tree \cite{De}. An extensive review of A/V classification methods can be found in \cite{miri2}. The A/V classification is never performed simultaneously with vessels segmentation as a 3-class classification task. Instead, either a manual or automatic segmentation is performed as a first step, a drawback being that any segmentation errors will be propagated to the A/V classification step.
The local features used to train the classifiers are based on contrast, intensities, derivatives or color features and the classifiers used include K-NN \cite{niemeijer} and linear discriminant analysis (LDA) \cite{dasht}.
Niemeijer et al. use structural information by aggregating the K-NN results for each branch of the vessel network \cite{niemeijer}.
Dashbozorg et al. correct the graph representing the vascular tree by applying rules derived from a priori knowledge of the retinal vasculature (for example, angles at bifurcation) \cite{dasht}.
The solution space of labeled graphs is efficiently explored by Estrada et al. by estimating graph likelihood model parameters such as color features, crossing rules, angles and topology \cite{estrada}.

Recently, more advanced machine learning methods have exploited recent advances in deep learning.
Convolutional Neural Networks (CNN) are particularly appropriate for image classification as its neurons are locally restricted to a portion of the image and the image content is highly locally correlated \cite{lecun,alex}. Promising results in various biomedical applications have been made by CNNs \cite{miccaicnn1}. For example, glaucoma detection \cite{glaucomacnn} and dark hemorrhages segmentation \cite{darkcnn} have been performed by training CNNs. Vessels segmentation using CNNs has been able to outperform the majority of previous state-of-the-art methods \cite{Fu2016, liskowski}. Recent works have proposed pixelwise CNN classification of vessels into arteries or veins after segmentation of the vessels \cite{ukbiobank, pulmonart}. We have recently ourselves proposed a simple CNN model that performs A/V classification from a pre-segmented vascular network \cite{fantin}. This preliminary study proposed an A/V classification method using a simple CNN that classified pixels into artery and vein. In those works, the inference speed is slow as each pixel is classified separately. An automatic segmentation of the vascular network is required before the CNN classification, which makes the classification performance dependent upon that of the segmentation.

In this paper, we propose a novel fully automatic A/V semantic segmentation method that performs the whole task of segmenting the vessels and classifying them as arteries or veins directly from the fundus images using a single encoder-decoder CNN model. The A/V segmentation method proposed in this work involves two major steps: \textbf{1)} initial classification into background, arteries or veins with a CNN; \textbf{2)} propagation of the CNN labeling output through the vascular network. The second step uses a graph representation to capture the global structural information of the vascular network. 

The first main contribution of this work is to propose a semantic segmentation method that simultaneously segments and classifies vessels into arteries and veins using deep-learning techniques. It outperforms many recent works, including several using deep learning techniques. The method we proposed is fast and easily scalable to any fundus image size. The CNN model proposed in this paper is designed to be very fast at inference time. 
Two more specific contributions are also worth mentioning: first, the originality of our training strategies, which include representing the image with six channels, augmenting the data using realistic principal component analysis (PCA) augmentation, and selecting training patches close to the vessels; second, the efficient scheme used to propagate the CNN's labeling through the vascular tree.

The standard procedure for calculating the AVR usually relies on the widest vessels, i.e. those closest to the OD, as described in \cite{niemeijer} and \cite{knudtson}. However, it is known that smaller arterioles are more affected than larger ones by high blood pressure. In that light, our second main contribution is to propose a novel global AVR measure that uses the automatically classified arterioles and venules in the entire field of view of the fundus image, as opposed to the standard local AVR measure that considers only the vessels close to the OD. Using this global AVR, we can detect significant changes not only between healthy and proliferative DR cases but also between healthy and moderate DR cases, which is not true using the local AVR.

The paper is organized as follows. In Section \ref{sec:methods}, we present the methodology for jointly segmenting and classifying the vessels in a fully automatic manner. In Section \ref{sec:results}, we present a comprehensive evaluation and discussion of our method, including comparisons with the state of the art both for the vessel segmentation and A/V classification tasks. We then use our joint segmentation and classification method to evaluate the changes in AVR for different DR grades and compare the proposed global AVR and standard local AVR measures. Section \ref{sec:conclusion} concludes this article and presents ongoing and future work.

\section{Methodology}\label{sec:methods} 

The A/V semantic segmentation method proposed in this paper involves two major steps: initial semantic segmentation using a CNN model, followed by propagation of the CNN output. In this propagation step, the vasculature is represented as a mathematical graph inside which the CNN labels are spread.

\subsection{Semantic segmentation with a convolutional neural network}
\begin{figure*}[!t]
\centering
\includegraphics[width=6in]{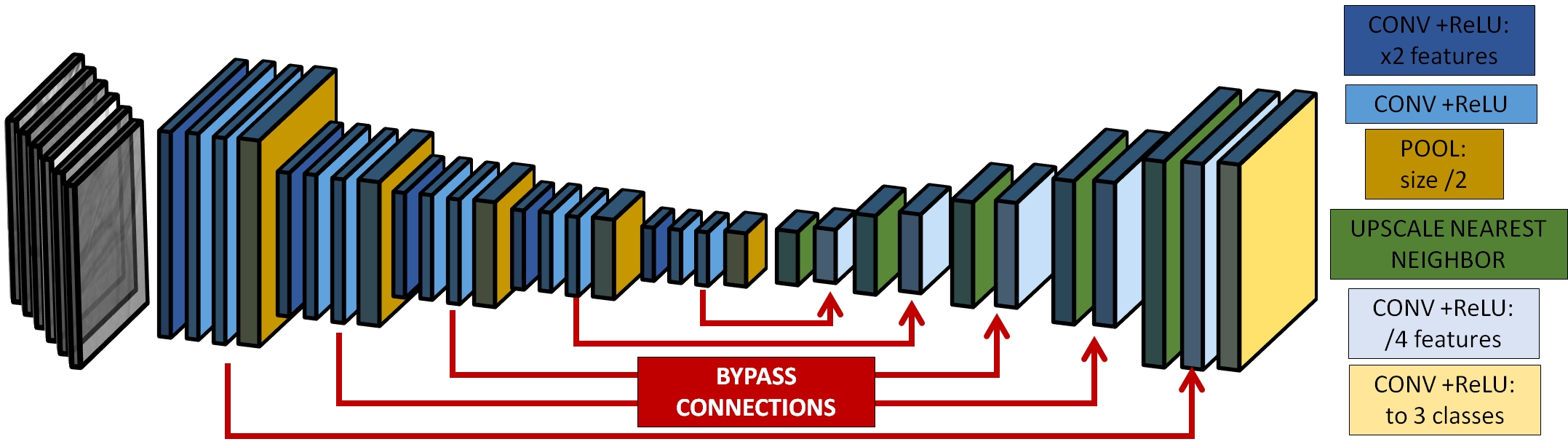} 
\caption{Proposed CNN model, using five encoding convolutional layers and five decoding convolutional layers along with bypass connections. The legend at right identifies the different types of layers.}
\label{fig_cnn}
\end{figure*}

The proposed encoder-decoder CNN model, shown in Fig. \ref{fig_cnn}, is comprised of encoding layers that encode the input into smaller and smaller vectors, and decoding layers that upsample the small vectors into a labeled segmented image. The proposed architecture is appropriate for generalization as it encodes the information into a smaller vector. 
We use the idea from U-Net \cite{miccaicnn1} of bypassing the encoding-decoding portion by reintroducing feature maps from each encoding layer to the corresponding decoding layer. In this way, some high-resolution boundaries are more likely to be preserved from the encoding part of the network. Our model is similar to the U-Net model but more flexible as it can take, for training or inference, input patches of any size, even the whole image at the same resolution. It outputs a labeled segmented image of the same size of the input as opposed to U-Net model that crops the images after each convolution. It allows fast inference as the model can take the whole image as input at the inference phase, thus avoiding the need to use the overlap-tile strategy.

\subsubsection{Preprocessing}
Fundus images can have nonuniform illumination issues due to small pupil size and over/under exposure during acquisition. These defects are visible in the whole image as low frequency artefacts. As the training input of the CNN is made of relatively small image patches, the CNN will not be able to correct these illumination issues by itself.
To correct the illumination, median filtering is applied with a kernel size equal to one tenth of the vertical field-of-view (FOV). The result of this operation is a good estimation of the low-frequency image illumination. These median-filtered channels $I^c_{med}$ are subtracted from the original RGB channels $I^c$. The standard deviation is normalized to a fixed value $\sigma_0$ by dividing the resulting image $I^c-I^c_{med}$ by its global standard deviation $\sigma_{I^c-I^c_{med}}$ and then multiplying by the fixed standard deviation value to obtain the three normalized RGB channels $I^c_{norm}$:

\begin{equation}
\label{eq0}
I^c_{norm} = \sigma_0\frac{I^c-I^c_{med}}{\sigma_{I^c-I^c_{med}}}+128
\end{equation}

Finally, these three normalized channels are concatenated to the original RGB channels of the fundus image, giving the input images composed of six channels.

\subsubsection{Layers description}

The convolutional layer $i$ at depth $n$ is composed of a local 3x3 convolution from the six-channels input patch or from the output of the previous layer,followed by a linear addition of a bias and finally transformed by a non-linear function. The rectified linear unit (ReLU) is used here to produce the output $\mathbf{l_i^{n}}$ as defined in Equation \ref{eq1}. The trainable parameters are the weights of all the convolution kernels $\mathbf{k_{ij}^n}$ and the corresponding biases $\mathbf{b_i^n}$.

\begin{equation}
\label{eq1}
\mathbf{l_i^n} = \max(0, \sum_j\mathbf{l_j^{n-1}}\ast\mathbf{k_{ij}^n}+\mathbf{b_i^n})
\end{equation}

Each encoding layer has 3 stacked convolutional layers (the first one of which increases the feature map size by a factor of two) followed by a max pooling layer. Each decoding layer has an upscaling layer, which performs nearest neighbor upscaling, followed by a convolutional layer, which reduces the number of feature maps by four. Each max pooling layer reduces the image size by a factor of two, while each upscaling layer increase it by a factor of two. The number of feature maps is 32 for the first encoding layer, 64 for the second, 128 for the third, 256 for the fourth and 512 for the fifth. After each upscaling layer, the feature maps are concatenated with the corresponding ones from the bypass connections, yielding twice as many feature maps. The number of feature maps is then reduced by four by the decoding convolutional layers. So the numbers of feature maps after each decoding convolutional layer are respectively: 256, 128, 64, 32, 16. At the end, a final convolutional layer reduces the feature maps from 16 to 3 classes (background, artery and vein) and the output of the model is a three-channel patch of same size as the input. Each channel of the output patch expresses the pixelwise probability of being either background, artery or vein. The final pixelwise classification label corresponds to the class whose probability is the highest.

\subsubsection{Training Strategies}
At the training stage, the proposed model is fed by six-channels image patches of size 128x128. The number of labeled fundus images is currently limited which make the training with the whole image not achievable.
We use Glorot initialization method to set up initially the weights and biases \cite{xavier}.
For each batch size $n_{batch}$, cross-entropy loss function is minimized and the resulting update from the ADAM stochastic gradient descent \cite{adam} are back propagated to the weights and biasesof the network.
One important strategy is how we select the training patches. The initial patches extracted are centered on the vessels. Then, we move randomly around the vessels and add other patches centered on the background but close to the vascular network. The CNN is forced in this manner to learn to distinguish the background from the vessels where this is most important, i.e. close to the vessels. Compared to simply taking patches randomly, the training data is more balanced in our case, with most of the patches containing pixels from the three classes,
thereby improving the convergence of the network. In areas far from the vascular network, the A/V classification accuracy is less critical as false positives in the background can be easily removed via morphological operations.

\subsubsection{Data Augmentation}
\begin{figure}[!t]
\centering
\includegraphics[width=3.3in]{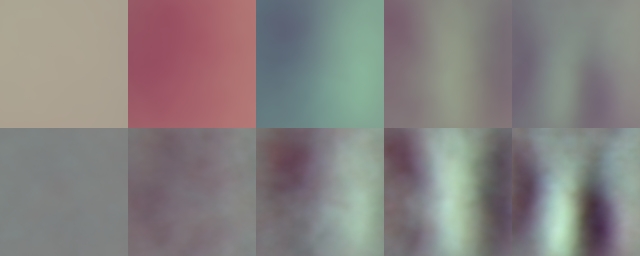} 
\caption{Intensity variability of training patches. From left to right : first five eigenvectors from PCA. Top row: the three standard RGB channels; Bottom row: the three normalized RGB channels.
}
\label{fig_eig1}
\end{figure}

\begin{figure}[!t]
\centering
\includegraphics[width=3.3in]{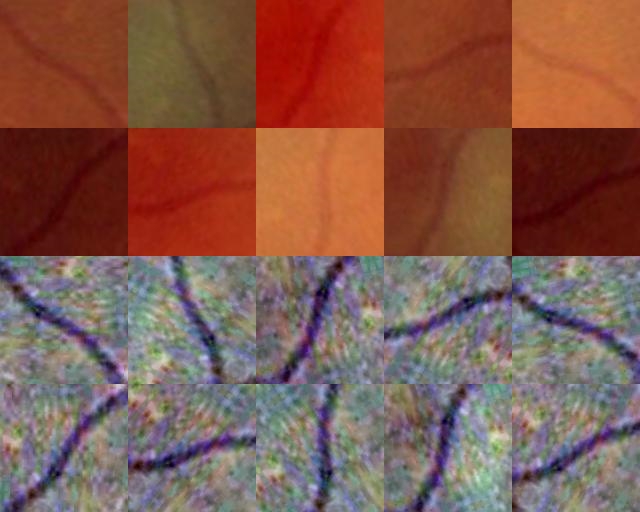} 
\caption{Example of 10 data augmentations of the same six-channel patch. Top two rows: standard RGB image patches; Bottom two rows: corresponding three-channel normalized image patches. Our data augmentation strategy covers a wide range of realistic colors.}
\label{fig_eig2}
\end{figure}

To avoid overfitting the model, data augmentation is performed at the training stage. This consists in artificially augmenting the training data and thus adding more variability to the training dataset. Geometric and iconic transformations are performed on each training patch. An important property of these transformations, especially in medical imaging, is that the synthetic image patches must be the most realistic possible. To achieve this, we consider only 2D rotations (by random angles) for geometric augmentation. As for iconic transformations, the illumination and background color differences across fundus images add a lot of variability in the intensities. This variability of pixel intensities is encapsulated in the original training dataset and can be analyzed using principal component analysis (PCA). To each original patch, we add a random fraction of an eigenvector chosen randomly among the five principal eigenvectors. The variability of the image intensities is captured by these eigenvectors (see Fig. \ref{fig_eig1}). As a result, the original patches are transformed randomly during the batch preparation, thereby augmenting the training data  while remaining within the range of what exists in fundus images.
We can see in Fig. \ref{fig_eig2} that the augmented patches cover a wide range of realistic choroid colors and vessels orientations observed in fundus images.

\subsection{Likelihood score propagation (LSP)}
The CNN model outputs a labeled map corresponding to the maximum at each pixel position of three probability maps (one for each class), denoted $p_{back}$ for the background, $p_{artery}$ for the arteries and $p_{vein}$ for the veins). 
For each positions $(x,y)$, a likelihood score $s(x,y)$ is defined as follows : if close to 1, the likelihood of being an artery is high; on the contrary, if close to 0, the likelihood of being a vein is high:

\begin{equation}
\label{eq17}
s(x,y)= \frac{p_{artery}(x,y)}{p_{artery}(x,y)+p_{vein}(x,y)}
\end{equation}

The patches used to train the CNN encode limited information about the global topology of the blood vessels, i.e. the connectivity of the vein/artery tree. Using a graph representation, we can efficiently propagate the initial CNN labeling through the vascular network. This propagation involves two steps: first, the vessel branches are extracted to form the graph representation. Then, the CNN scores are propagated using the minimum spanning tree of the graph.

\subsubsection{Extraction of Vessel Branches}
\begin{figure}[!t]
\centering
\includegraphics[width=3.3in]{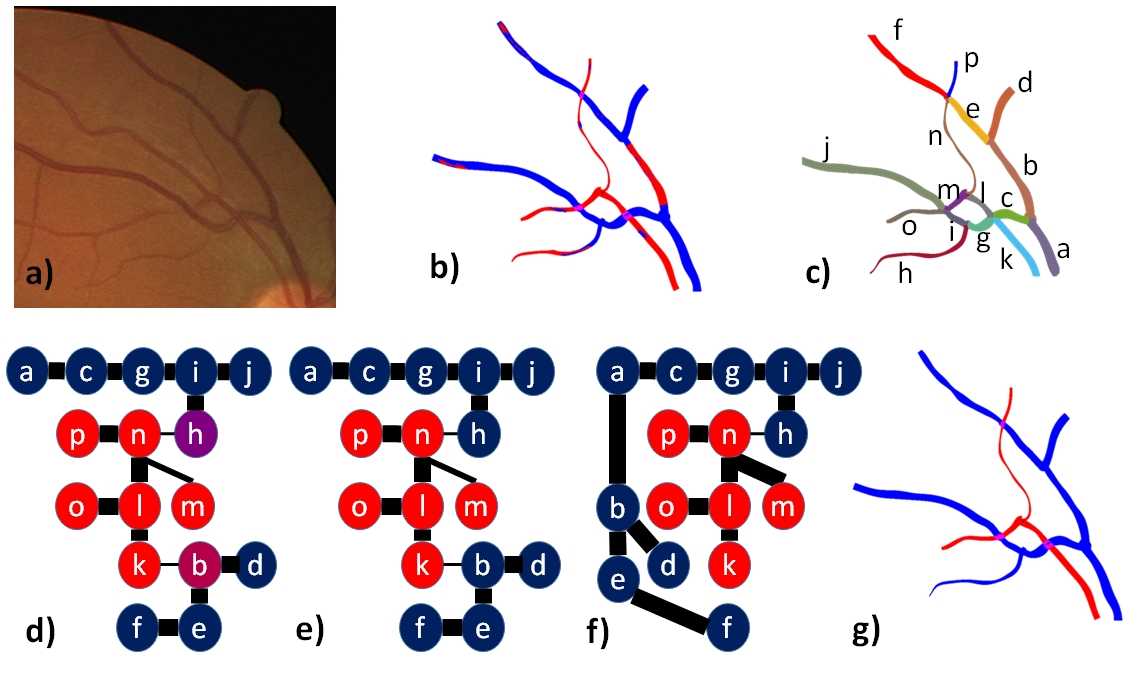}
\caption{Likelihood score propagation (LSP): a) fundus image; b) CNN output ; c) connected components; d) corresponding minimum spanning tree with node scores (from blue for veins to red for arteries) and edge costs (high cost and low propagation = thin line, low cost and high propagation = large line); e) minimum spanning tree after the first iteration; f) minimum spanning tree after the second iteration g) final classification of vessels. }
\label{fig_propa}
\end{figure}

The vessel segmentation contains all the pixels not labeled by the CNN as background (see Fig. \ref{fig_propa} b)). This vessel segmentation is first skeletonized using Zhang-Suen method \cite{zhangthinning}.
Then, the connected components of the vessel tree (see Fig. \ref{fig_propa} c)) are obtained with the classical two-pass algorithm after removing bifurcations and crossings. The bifurcations and crossings are identified with hit-and-miss morphological operations. The set of $N_i$ pixels belonging to a given connected component forms a branch, called $b_i$. The two endpoints $\mathbf{X}^i_1$ and $\mathbf{X}^i_2$ of each branch $b_i$ and the orientations $\alpha^i_1$ and $\alpha^i_2$ at these points are calculated and will be used for the graph representation.

\subsubsection{Graph Representation}
The retinal vasculature is represented mathematically as a connected, undirected graph $G=(V,E)$ with a cost function $c:E\mapsto \mathbb{R}$. The nodes $V$ of this graph are the vessel branches $b_i$. Each node is associated to the branch's likelihood of being an artery, $s_i$. This branch score $s_i$ is calculated from the pixel scores $s(x,y)$ previously obtained from the CNN classification and re-centered around zero:

\begin{equation}
\label{eq2}
s_i = s(b_i) =  \frac{1}{N_i}\sum_{(x,y)\in b_i}s(x,y) -0.5
\end{equation}

The weight associated to each edge of the graph expresses the cost $c$ of linking two graph nodes (i.e. vessel branches). If the cost is low, it means the likelihood is high that two branches $b_i$ and $b_j$ are connected in the retinal vascular tree. This cost $c$ is defined as the sum of a position cost $c_{pos}$ and a label cost $c_{lab}$. The weight of the label cost relative to the position cost is controlled by two parameters $\sigma_{pos}$ and $\sigma_{lab}$. The position cost is defined as the minimal distance between the two endpoints of each branch to which we add an extra cost $\lambda_{angle}$ on angles in order to favor the connection of collinear branches. The label cost $c_{lab}$ is equal to the difference between the two branches' scores $s_i$ and $s_j$: 

\begin{multline}
\label{eq3}
c_{pos}(b_i, b_j) = \frac{1}{\sigma_{pos}}\min_{k\in\{1,2\}, l\in\{1,2\}}\|\mathbf{X}^i_k-\mathbf{X}^j_l\| \\ +\lambda_{angle}|\alpha^i_k-\alpha^j_l|
\end{multline}

\begin{equation}
\label{eq5}
c_{lab}(b_i, b_j) =  \frac{1}{\sigma_{lab}}|s_i -s_j|
\end{equation}

\subsubsection{Score Propagation in the Minimum Spanning Tree}

The idea is to propagate the scores $s_i$ from one node to another according to the cost of the edge connecting those two nodes. If the cost is low, then the propagation should be greater, 
and vice-versa. To implement this process, each node $b_i$ receives a likelihood score from another node $b_j$. This score received by $b_i$ is exponentially attenuated by the position cost $c_{pos}(b_i,b_j)$ between the two nodes. 
But instead of calculating, for a given node $b_i$, the likelihood scores from all $b_j\in V$ attenuated by the costs $c_{pos}(b_i, b_j)$, an efficient method of score aggregation used for stereo correspondence \cite{yang} is adapted for our application. 
The graph is first simplified into its minimum spanning tree, obtained with the Prim algorithm \cite{prim} (see Fig. \ref{fig_propa} d) for an example). In this minimum spanning tree, the sum of the costs along the path connecting two nodes should be a good approximation of the total cost between these two nodes in the original graph. 
The tree is then traversed in post-order from the leaves up to the root, and each initial score $s_i$ is updated to an intermediate score denoted $s_i^\uparrow$.
To do this, each parent $b_i$ receives from its children $b_j$ their intermediate scores $s_j^\uparrow$ exponentially attenuated by the position costs $c_{pos}(b_i, b_j)$, and these are added to the parent's current score. We use the following notation $b_i=P(b_j)$ meaning that the parent of node $b_j$ is $b_i$.
The intermediate scores of the leaves are simply their initial scores, since they have no children. The amount of attenuation for a specific cost is controlled by $\sigma_{prop}$:

\begin{equation}
\label{eq8}
s_i^\uparrow = s^\uparrow(b_i) =  s_i + \sum_{b_i\in V, P(b_j)=b_i}\exp(\frac{c_{pos}(b_i, b_j)}{\sigma_{prop}})s_j^\uparrow
\end{equation}

The remaining likelihood scores to propagate are those coming from sibling nodes. Since upward propagation has already occurred, it is possible to calculate the remaining propagation coming from the siblings. The tree must now be traversed in pre-order from the root down to the leaves. 
The intermediate scores $s_i^\uparrow$ are updated to the final scores $s_i^{fin}$. 
For the root, as it has no parent, the final score is simply equal to its intermediate score. Each parent $b_i$ then transmits to each of its children $b_j$ the likelihood scores $s_i^{fin}$ (that now contain the aggregation of scores from all subtrees) minus the likelihood score already sent by that child, resulting in the final score for each child $s_j^{fin}$ (see the resulting tree in Fig. \ref{fig_propa} e):

\begin{multline}
\label{eq9}
s_j^{fin} =  s_j^\uparrow + \exp(\frac{c_{pos}(b_i, b_j)}{\sigma_{prop}})[s_i^{fin} \\ -\exp(\frac{c_{pos}(b_i, b_j)}{\sigma_{prop}})s_j^\uparrow]
\end{multline}

This entire procedure is repeated until convergence or for a sufficient number of iterations. This allows some nodes to be reconnected properly in the minimum spanning tree so that they can receive the appropriate score propagation (for example the node b in Fig. \ref{fig_propa} f)). Experimentally, we found that two iterations were usually sufficient to reach convergence. 

Finally, the labeling of the vessel segmentation is updated according to the branches' final scores $s_i^{fin}$. The branch is labeled as a vein if $s_i^{fin}$ is negative and as an artery if $s_i^{fin}$ is positive (see Fig. \ref{fig_propa} g)).

\section{Experiments and discussion}\label{sec:results} 

We present in this section the experiments conducted and compare our results with the state of the art by reporting segmentation results as well as classification results.

\subsection{Parameters and training strategies selection}

10-fold cross validation is applied to tune the parameters for the LSP ($\sigma_{like}$, $\sigma_{prop}$, $\sigma_{pos}$, $\lambda_{angle}$). $\sigma_{like}$, $\sigma_{prop}$, $\sigma_{pos}$ and $\lambda_{angle}$ were set respectively to 0.1, 10.0 and 100.0. To do so, we first selected the best set of parameters for the CNN model. With these parameters fixed, we then performed the cross validation to choose the best set of parameters for LSP.
For the preprocessing step, we tested different image contrast enhancement methods including CLAHE, top/bottom hat and median filtering, the latter proving to be the best method. 
For the CNN model, we tested both nearest neighbor upscaling and deconvolution for the upscale layers. The former approach decreased the number of artifacts in the background and was thus deemed better than the deconvolution operation. The batch size $n_{batch}$ was set to 128.
For each model learned, we stopped the learning and used the model trained after 50 epochs (meaning that the CNN has seen the training dataset 50 times, each time altered through augmentation strategies).
The number of layers was chosen experimentally. We determined that increasing the number of convolutional layers before pooling decreases the false positive rate. Stacking three 3x3 convolutional layers is equivalent to calculating 7x7 convolutions. Hence we may surmise that the number of stacked convolutional layers is linked to the size of the objects that the model must learn. Vessel size (width) is more stable than the size of the structures in the background. This can explain the decrease in false detections in the background observed when increasing the number of stacked convolutional layers.
The data augmentation strategy, that applies random rotations and PCA-based alterations of intensities on the training exemplar,  reduces the error rate by 1 percentage point.

\subsection{Data}
The public datasets DRIVE \cite{staal} and MESSIDOR \cite{decenciere_feedback_2014} were used to train and test our CNN model to favor comparison and replicability of the methodology proposed in this work. The DRIVE dataset consists of 40 fundus images divided into 20 training images and 20 test images. For the vessel segmentation ground truth, the training set was annotated by one expert, giving one training annotation set, while the test set was annotated by two experts, giving two test annotation sets. It is worth noting that for the test annotation sets, the first expert is the same one who annotated the training data, while the second expert is independent from the training dataset \cite{staal}. 
Regarding the A/V classification, we used two gold standards: ALL-DRIVE in which all vessel pixels are labeled as artery or vein, \cite{rite2013}; and CT-DRIVE, in which only the centerlines of the vessels of diameter greater than 3 pixels are labeled as artery or veins\cite{dasht}.
For the MESSIDOR dataset that contains 1200 fundus images, a clinical expert has labeled the vessels into arteries or veins in 100 fundus images. 70 images from this labelled subset are used for the training phase and the other 30 for testing.

\subsection{Performance results of whole system} 

To assess our results, we report the accuracy measure of the three-class classification, which corresponds to the proportion of pixels inside the FOV mask that belong to the correct classes in the output map. 

The performance of the vessel segmentation and artery/vein classification tasks 
are evaluated using sensitivity (true positive rate) and specificity (true negative rate) measures obtained by thresholding the probability maps, the positive classes being either vessels or arteries: 

\begin{equation}
\label{eq17}
TPR(z)=\frac{1}{N_{pc}} \sum_i^{N_{pc}}p_{pc}(i)\geq z
\end{equation}

\begin{equation}
\label{eq17}
TNR(z)=\frac{1}{N_{nc}} \sum_i^{N_{nc}}p_{nc}(i)<z
\end{equation}

where $N_{pc}$ is the number of pixels actually belonging to the positive class inside the FOV mask, $p_{pc}(i)$ is the probability of a pixel to belong to the positive class according to the output of our method, $N_{nc}$ is the number of pixels actually belonging to the negative class, and $p_{nc}(i)=1-p_{pc}$ is the probability of a pixel to belong to the negative class. 
For vessel segmentation, the positive and negative classes are vessel and background respectively; for A/V classification, they are artery and vein respectively.
As another performance metric, we use the AUC criterion, which represents the area under the receiver operating characteristic (ROC) curve. The ROC curve is constructed by changing the threshold applied to the probability map and plotting the sensitivity/specificity pairs for all threshold values in the range [0, 1]. The AUC value is then calculated using the trapezoidal rule \cite{delong}.

We define the vessel probability map $p_{vessels}$ with the following equation:

\begin{equation}
\label{eq17}
p_{vessel}= \frac{\max(p_{artery}, p_{vein})}{p_{back}}
\end{equation}

We report the results in Table \ref{table1} and \ref{table2} for different versions of our method: \textit{CNN 3D} corresponds to the model trained with the original images, \textit{CNN 3D enhanced} to the model trained with the enhanced images, \textit{CNN 6D} to the model trained with the 6D images (original + enhanced images), \textit{CNN 6D BP} to the model with addition of bypass connections, and finally \textit{CNN 6D BP + LSP} is the CNN model followed by LSP propagation. We report the AUC for vessel segmentation, the sensitivity, specificity and accuracy for A/V classification, and the overall accuracy, i.e. for the classification of all image pixels into the three classes. 

\begin{table}[!t]\caption{Results for different versions of the method on DRIVE \label{table1}}
\centering
\resizebox{0.46\textwidth}{!}{%
\begin{tabularx}{0.50\textwidth}{c *{6}{s}}
\toprule
\multirow{2}{*}{Method}   & Vessels &   \multicolumn{3}{c}{Artery Vein} & All\\
\cmidrule{3-5}
& AUC & Sens. & Spec. & Acc. & Acc.\\
\midrule
CNN 3D  &  0.961 &  79.5\% & 78.2\% & 78.8\% &  93.90\% \\
CNN 3D enhanced & 0.968 & 80.0\%  & 85.6\% & 83.0\% & 94.31\%\\
CNN 6D & 0.968 & 80.9\% & 86.4\% & 83.8\% &94.34\% \\
CNN 6D BP & 0.972 & 81.6\% & 86.2\% & 84.2\% &94.59\% \\
CNN 6D BP + LSP & 0.972 & 86.3\%& 86.6\% & 86.5\% & 94.93\% \\ 
\bottomrule
\end{tabularx}}
\end{table}

\begin{table}[!t]\caption{Results for different versions of the method on MESSIDOR \label{table2}}
\centering
\resizebox{0.46\textwidth}{!}{%
\begin{tabularx}{0.50\textwidth}{c *{6}{s}}
\toprule
\multirow{2}{*}{Method}   & Vessels &   \multicolumn{3}{c}{Artery Vein} & All\\
\cmidrule{3-5}
& AUC & Sens. & Spec. & Acc. & Acc.\\
\midrule
CNN 3D  &  0.954 &  86.7\% & 86.3\% & 86.5\% &  94.8\% \\
CNN 3D enhanced & 0.976 & 87.0\%  & 92.4\% & 90.2\% & 96.0\%\\
CNN 6D & 0.977 & 89.4\% & 90.9\% & 90.3\% &96.1\% \\
CNN 6D BP & 0.982 & 88.0\% & 92.1\% & 90.4\% &96.3\% \\
CNN 6D BP + LSP & 0.982 & 95.3\%& 90.4\% & 92.4\% & 96.4\% \\ 
\bottomrule
\end{tabularx}}
\end{table}

We can see that using the 6D input is better than using only three channels, whether the enhanced or original ones. The enhanced channels mostly improve the vessel segmentation accuracy, while including the original image channels mostly improves A/V classification. 
In this respect, we note that although the contrast enhancement method we use corrects illumination differences at the whole image scale, the CNN model can learn the enhancement only at the local patch scale. 

\begin{figure}[!t]
\centering
\includegraphics[width=3.3in]{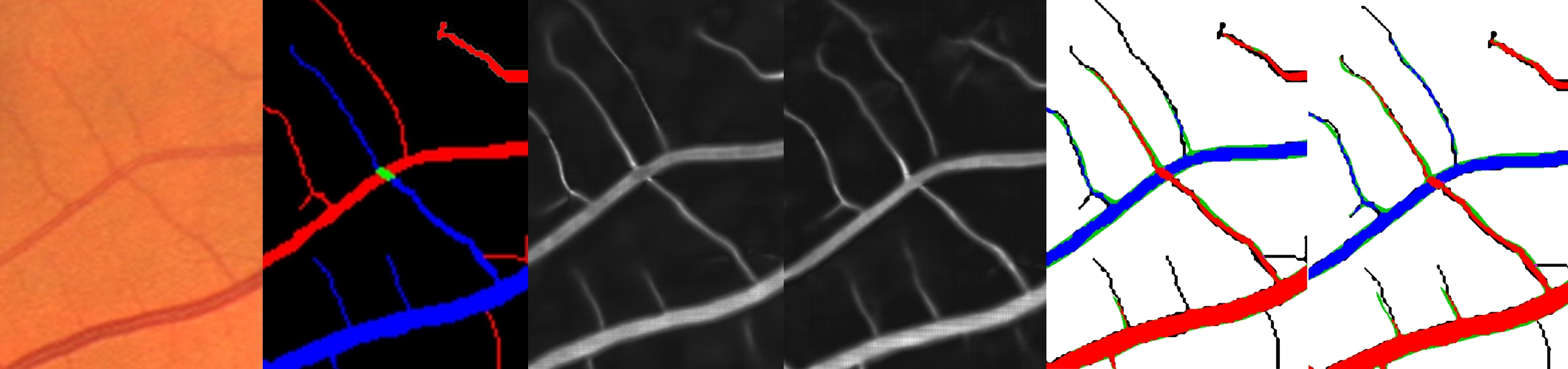}
\caption{Bypass connections improve segmentation of small vessels. From left to right: original image; ground truth; probability vessel map without BPs; vessels map with BPs; error map without BPs; and error map with BPs. In the error maps: blue pixels are true positive veins, red pixels are true positive arteries, black pixels are false negatives, green pixels are false positives and white pixels are true negative background pixels.
 }
\label{fig_BP}
\end{figure}

The bypass connections (BPs), which are the main idea of the U-Net model \cite{miccaicnn1}, mainly improve the segmentation performance. Reintroducing the high resolution feature maps 
to the decoder portion of the CNN model refines the boundaries of large vessels and the segmentation of small vessels (see Fig. \ref{fig_BP}). Inside the vessels, however, where the problem is to distinguish arteries from veins and where gradients are low, the high resolution information is less relevant. This can explain why the BPs do not improve A/V classification performance as much as the segmentation performance.

Finally, the LSP method, which propagates the vessel labels throughout the segmented vascular network, improves the A/V classification accuracy by 2 percentage points.

\begin{figure}[!t]
\centering
\includegraphics[width=1.8in]{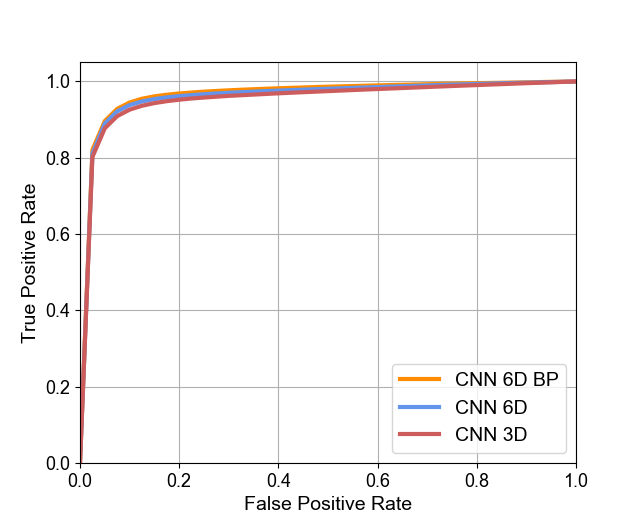}\includegraphics[width=1.8in]{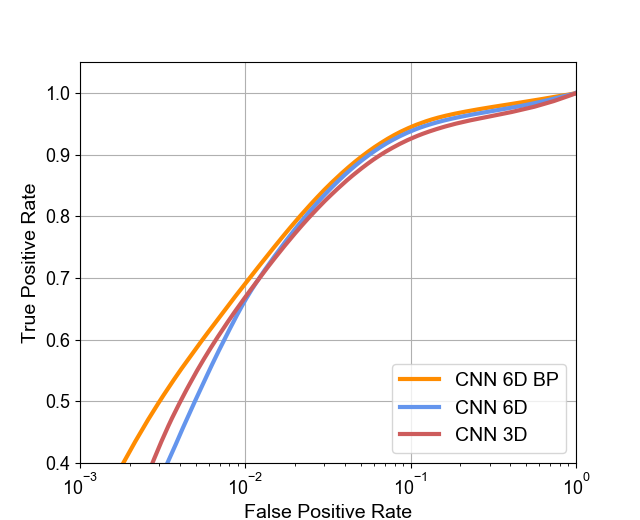}
\caption{Vessels segmentation on DRIVE: ROC curves.}
\label{fig_roc1}
\end{figure}

\begin{figure}[!t]
\centering
\includegraphics[width=1.8in]{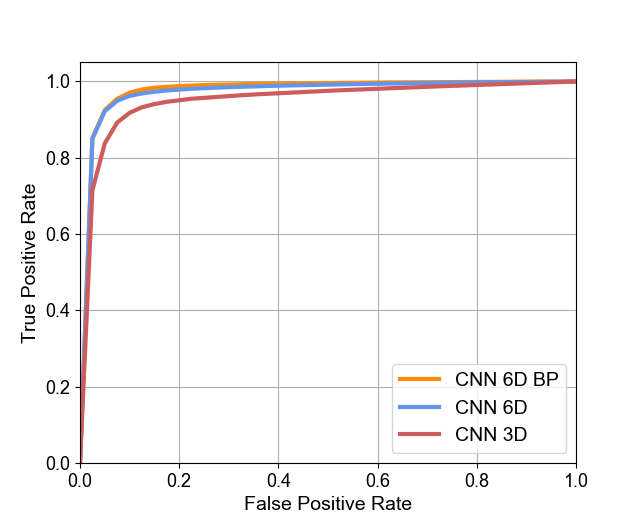}\includegraphics[width=1.8in]{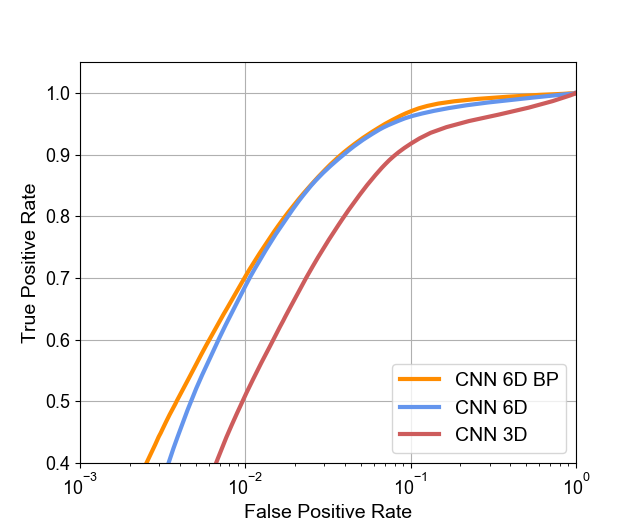}
\caption{Vessels segmentation on MESSIDOR: ROC curves.}
\label{fig_roc2}
\end{figure}

\begin{figure}[!t]
\centering
\includegraphics[width=1.8in]{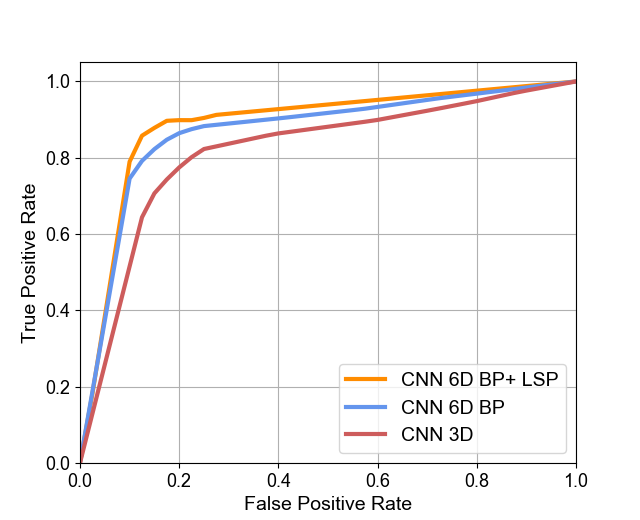}\includegraphics[width=1.8in]{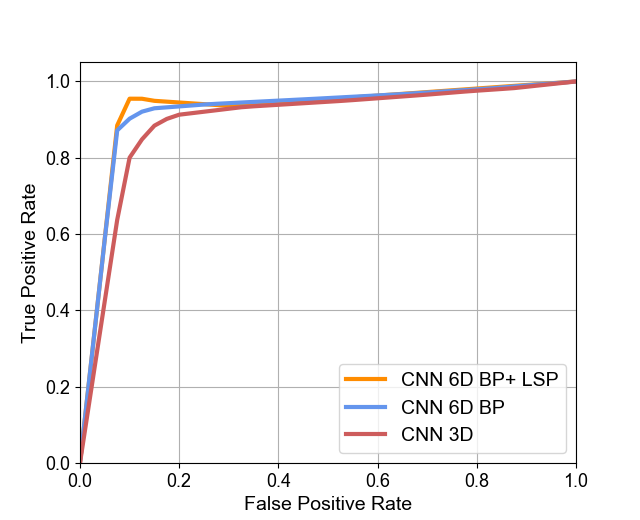}
\caption{A/V classification performance on DRIVE (left) and MESSIDOR(right): ROC curves.}
\label{fig_rocav1}
\end{figure}

The ROC curves for vessel segmentation are given in Fig. \ref{fig_roc1} and Fig. \ref{fig_roc2}, while the ROC curves for A/V classification are given in Fig. \ref{fig_rocav1}. We can see that LSP improves performance mostly near the equal error rate. Indeed, the LSP step will be able to correct mistakes in the CNN's prediction as long as the error level is reasonable; otherwise the LSP is likely to propagate errors.

\begin{figure}[!t]
\centering
\includegraphics[width=1.8in]{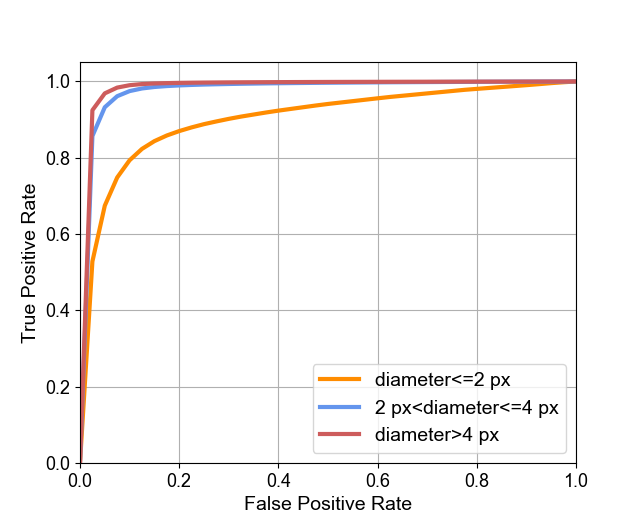}\includegraphics[width=1.8in]{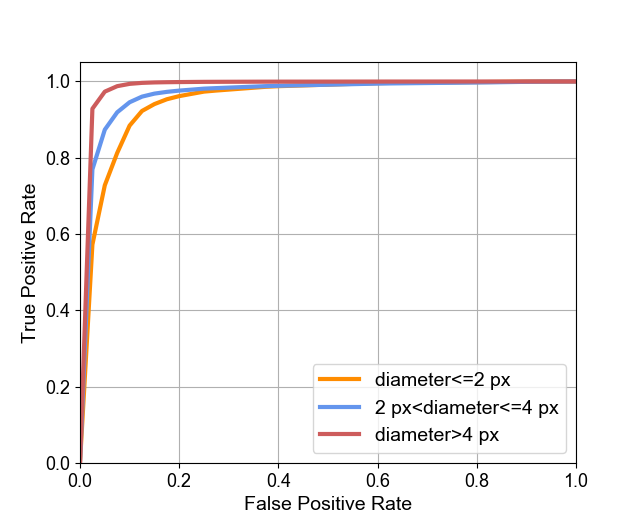}
\caption{Vessels segmentation on DRIVE (left) and MESSIDOR (right): the AUC is lower for smaller vessels.}
\label{fig_rocdiam1}
\end{figure}

We also analyzed our system's performance with respect to the vessels' diameters. Table \ref{tablestats} provides statistics on vessel diameters in the two image datasets. Fig. \ref{fig_rocdiam1} and Table \ref{tablesegdiam} show the ROC curves and performance results for vessel segmentation. Fig. \ref{fig_rocdiam1} and Tables \ref{tableavdiam1} and \ref{tableavdiam2} provide the ROC curves and performance results for A/V classification.
As expected, performance is lowest on the smallest vessels (inferior to 2 pixels) for both the segmentation and A/V classification tasks.

\begin{table}[!t]\caption{Statistics on vessel diameters \label{tablestats}}
\centering
\begin{tabularx}{0.48\textwidth}{c *{4}{s}}
\toprule
\multirow{2}{*}{Dataset}   & \multicolumn{3}{c}{\% of vessels of diameter (in pixels)}\\
\cmidrule{2-4}
& $<$2 & $\geq$2 and $<$4& $\geq$4\\
\midrule
DRIVE  &  19\% &  33\% & 47\%  \\
MESSIDOR & 1\% & 46\%  & 53\%\\
\bottomrule
\end{tabularx}
\end{table}

\begin{table}[!t]\caption{Segmentation results for different vessel diameters \label{tablesegdiam}}
\centering
\begin{tabularx}{0.48\textwidth}{c *{4}{s}}
\toprule
\multirow{2}{*}{Dataset}   & \multicolumn{3}{c}{AUC for diameter (in pixels)}\\
\cmidrule{2-4}
& $<$2 & $\geq$2 and $<$4& $\geq$4\\
\midrule
DRIVE  &  0.909 &  0.985 & 0.992  \\
MESSIDOR & 0.952 & 0.971  & 0.992\\
\bottomrule
\end{tabularx}
\end{table}

We can see in Tables \ref{tableavdiam1} and \ref{tableavdiam2} that the LSP stage mostly increases the accuracy for very small vessels, i.e. those between 1 and 4 pixels in diameter.
The more global structural information introduced by the LSP stage helps to improve performance where only local information is not sufficient such as on the smallest vessels.

\begin{figure}[!t]
\centering
\includegraphics[width=1.8in]{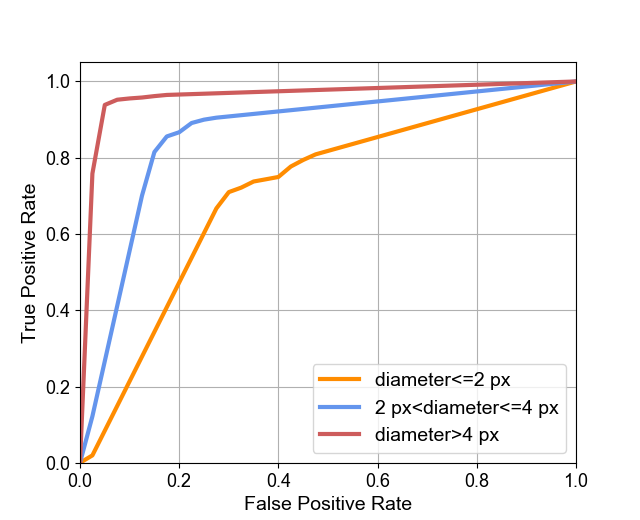}\includegraphics[width=1.8in]{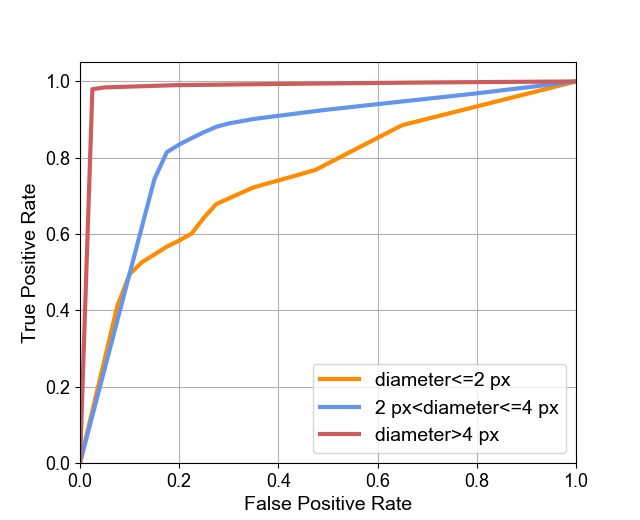}
\caption{A/V classification on DRIVE and MESSIDOR (right): The AUC is lower for smaller vessels.}
\label{fig_rocdiam2}
\end{figure}

\begin{table}[!t]\caption{A/V Classification results on DRIVE for different vessel diameters \label{tableavdiam1}}
\centering
\resizebox{0.5\textwidth}{!}{%
\begin{tabularx}{0.55\textwidth}{c *{4}{s}}
\toprule
\multirow{2}{*}{Method}   & \multicolumn{3}{c}{Accuracy for diameter (in pixels)}\\
\cmidrule{2-4}
& $<$2 & $\geq$2 and $<$4 & $\geq$4\\
\midrule
CNN 6D BP  &  67.5\% &  80.1\%  & 93.0\%  \\
CNN 6D BP + LSP  &  70.5\% &  84.2\% & 94.6\%  \\
\bottomrule
\end{tabularx}}
\end{table}

\begin{table}[!t]\caption{A/V Classification results on MESSIDOR for different vessel diameters \label{tableavdiam2}}
\centering
\resizebox{0.5\textwidth}{!}{%
\begin{tabularx}{0.55\textwidth}{c *{4}{s}}
\toprule
\multirow{2}{*}{Method}   & \multicolumn{3}{c}{Accuracy for diameter (in pixels)}\\
\cmidrule{2-4}
& $<$2 & $\geq$2 and $<$4& $\geq$4\\
\midrule
CNN 6D BP  & 70.2\% &  82.0\% & 97.8\%  \\
CNN 6D BP + LSP  &  73.5\% &  85.1\% & 98.3\%  \\
\bottomrule
\end{tabularx}}
\end{table}

\begin{figure}[!t]
\centering
\includegraphics[width=3.3in]{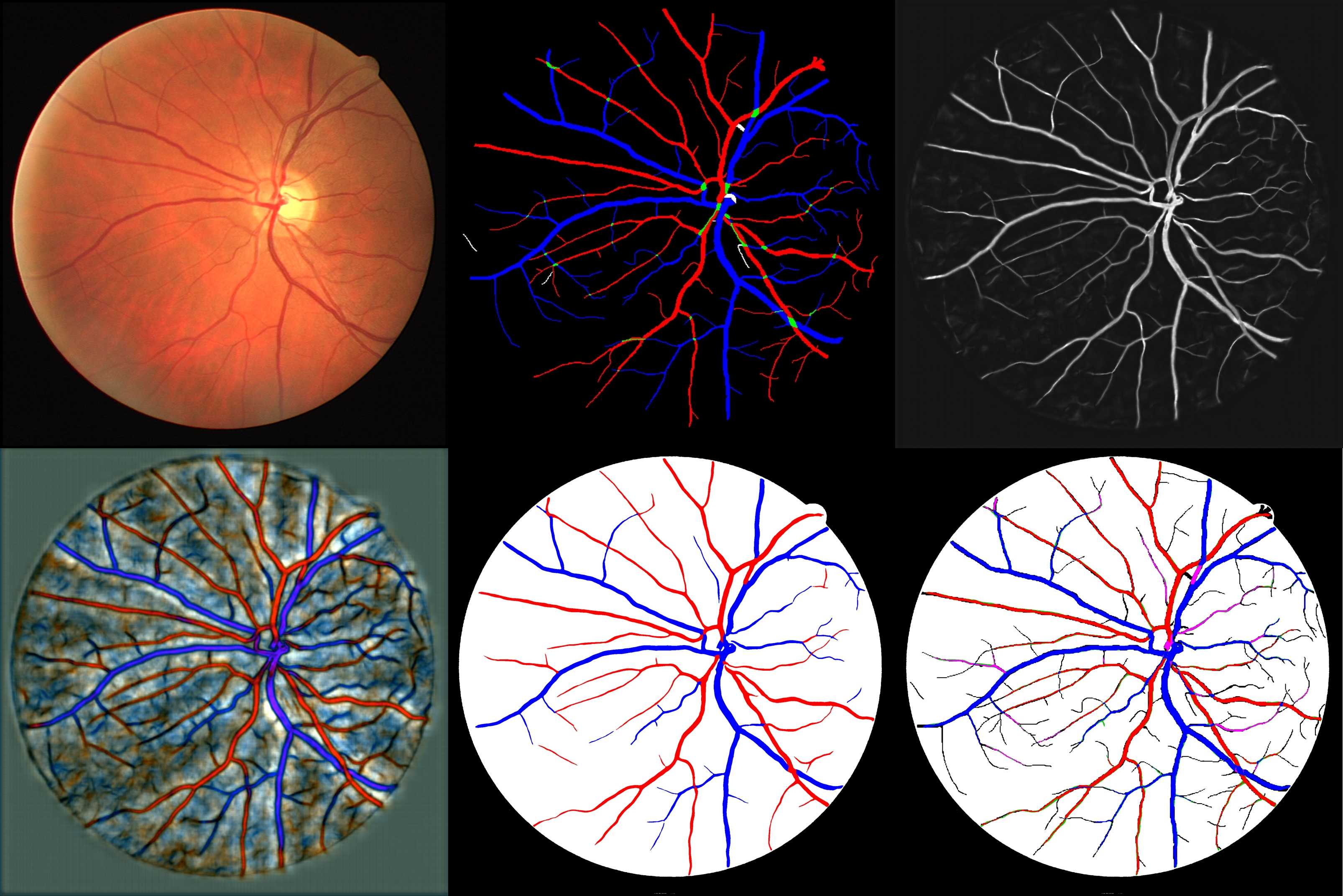}
\caption{Classification results with CNN 6D BP + LSP on 4th test image from DRIVE (Acc. = 95.4\%). Top row from left to right: 1) original image, 2) ground truth, 3) Vessel probability map. Bottom row from left to right: 1) CNN output probability map (blue: veins; red: arteries; gray: background), 2) hard classification, 3) error map (blue: true veins; red: true arteries; black: false negatives; green: false positives; white: true background).
 }
\label{fig_example1}
\end{figure}

\begin{figure}[!t]
\centering
\includegraphics[width=3.3in]{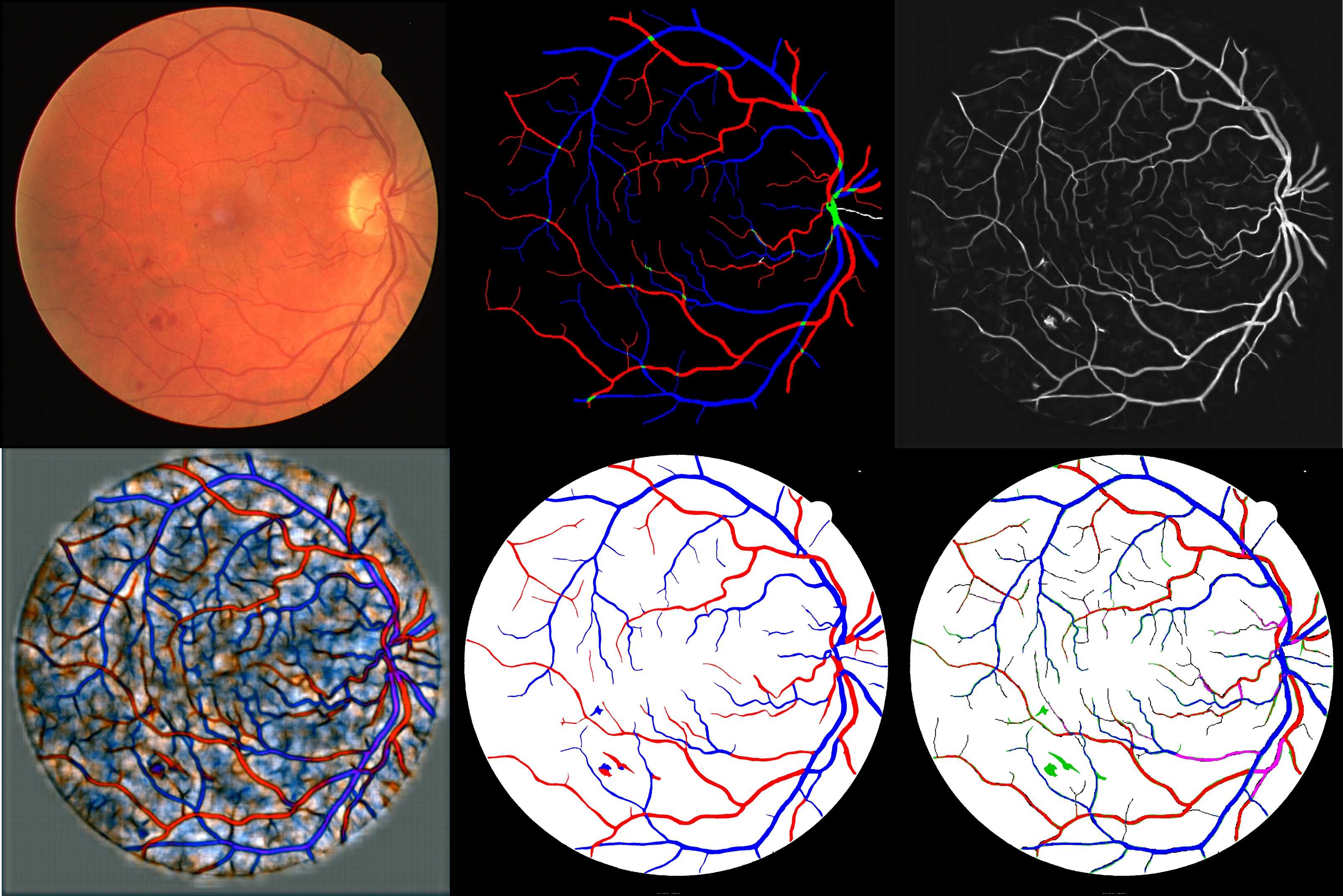}
\caption{Classification results with CNN 6D BP + LSP on 14th test image from DRIVE (Acc. = 95.3\%). Top row from left to right: 1) original image, 2) ground truth, 3) Vessel probability map. Bottom row from left to right: 1) CNN output probability map (blue: veins; red: arteries; gray: background), 2) hard classification, 3) error map (blue: true veins; red: true arteries; black: false negatives; green: false positives; white: true background).
}
\label{fig_example2}
\end{figure}

Analysis of the false negatives and false positives in the fundus images can help to better understanding the remaining errors. In Fig. \ref{fig_example1} and Fig. \ref{fig_example2}, two examples of outputs are given.
The false negatives are mainly small vessels that are not detected at all or are offset from their ground-truth locations. In Fig. \ref{fig_example2}, some hemorrhages are also detected as vessels.
In the vessels wider than 4 pixels, the false positives are concentrated at the borders of those vessels. 
In the vessels narrower than 4 pixels, the false positives are mainly due to their being offset from their ground-truth locations. Misclassification of arteries as veins or vice-versa is mainly concentrated on vessels narrower than 4 pixels. 
There are also some arteries and veins interlaced or very close to each other, also causing misclassification.

\begin{figure}[!t]
\centering
\includegraphics[width=2.8in]{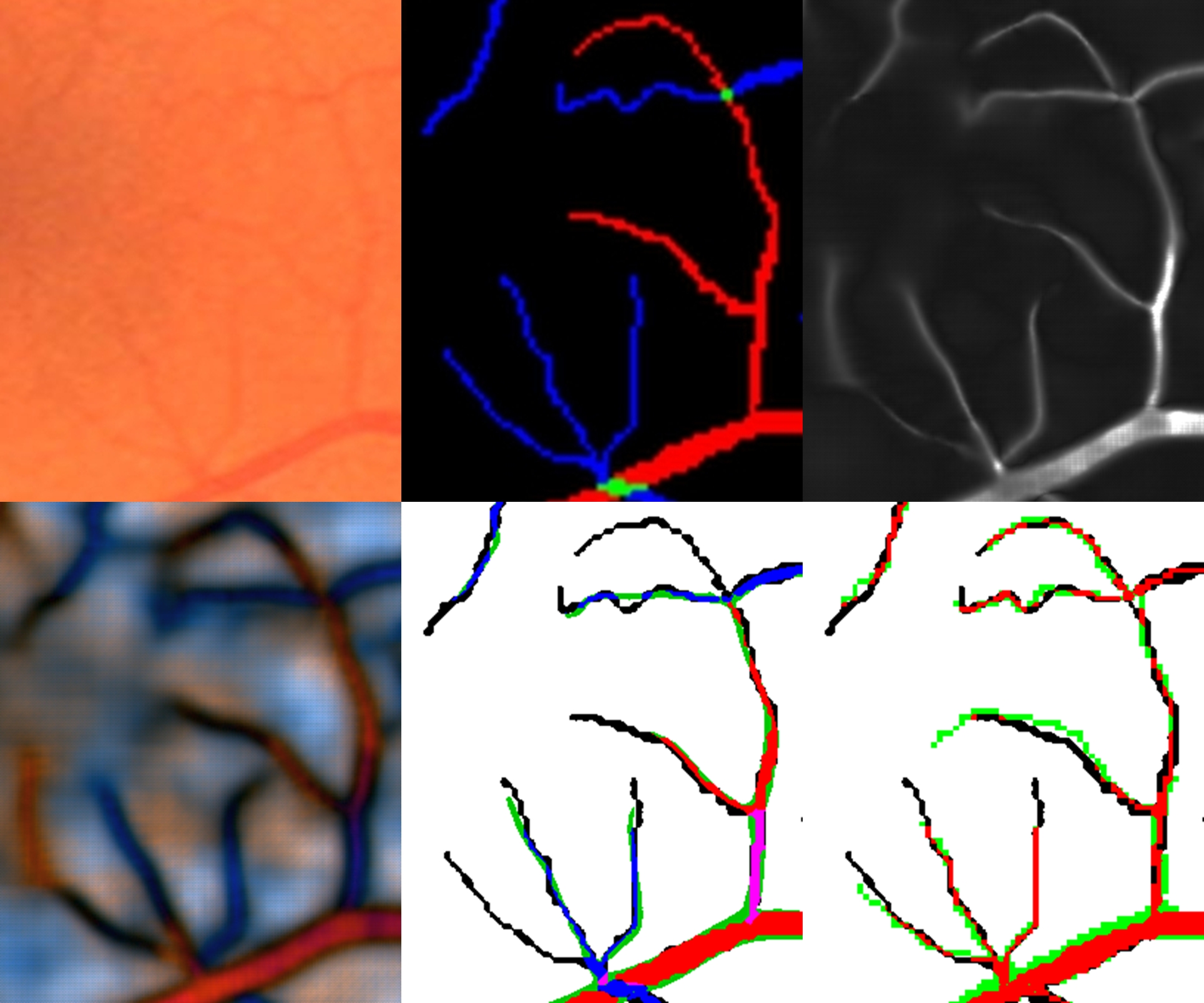}
\caption{Comparison of errors with a human expert. Top row from left to right: 1) original image, 2) ground truth, 3) Vessel probability map. Bottom row from left to right: 1) CNN output probability map (blue: veins; red: arteries; gray: background), 2) corresponding error map (blue: true veins; red: true arteries; black: false negatives; green: false positives; white: true background), 3) error map of the second expert (red: true vessels; black: false negatives; green: false positives; white: true background). }
\label{fig_refst1}
\end{figure}

\begin{figure}[!t]
\centering
\includegraphics[width=2.8in]{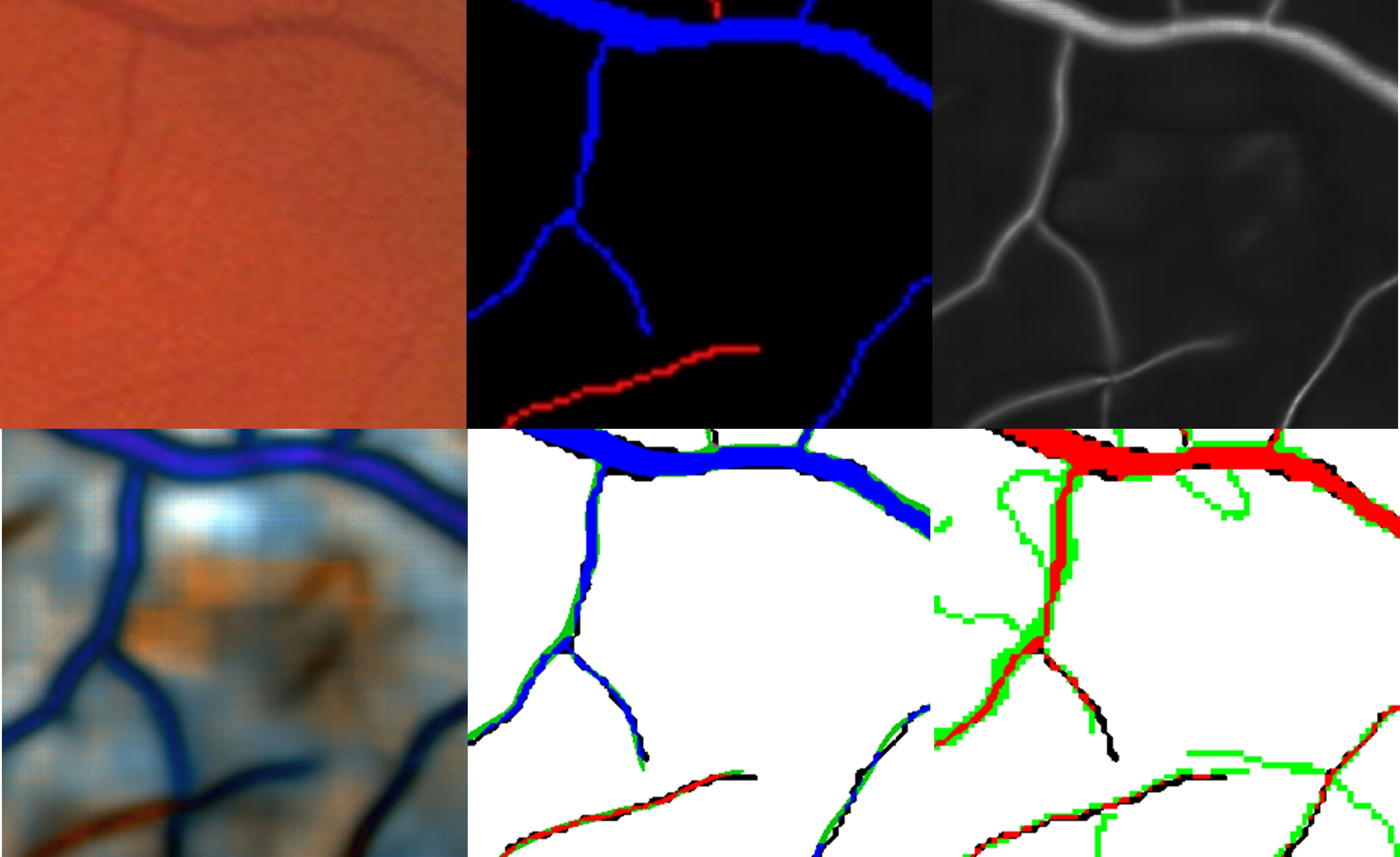}
\caption{Comparison of errors with a human expert. Top row from left to right: 1) original image, 2) ground truth, 3) Vessel probability map. Bottom row from left to right: 1) CNN output probability map (blue: veins; red: arteries; gray: background), 2) corresponding error map  (blue: true veins; red: true arteries; black: false negatives; green: false positives; white: true background), 3) error map of the second expert (red: true vessels; black: false negatives; green: false positives; white: true background). }
\label{fig_refst2}
\end{figure}

\begin{figure}[!t]
\centering
\includegraphics[width=3.0in]{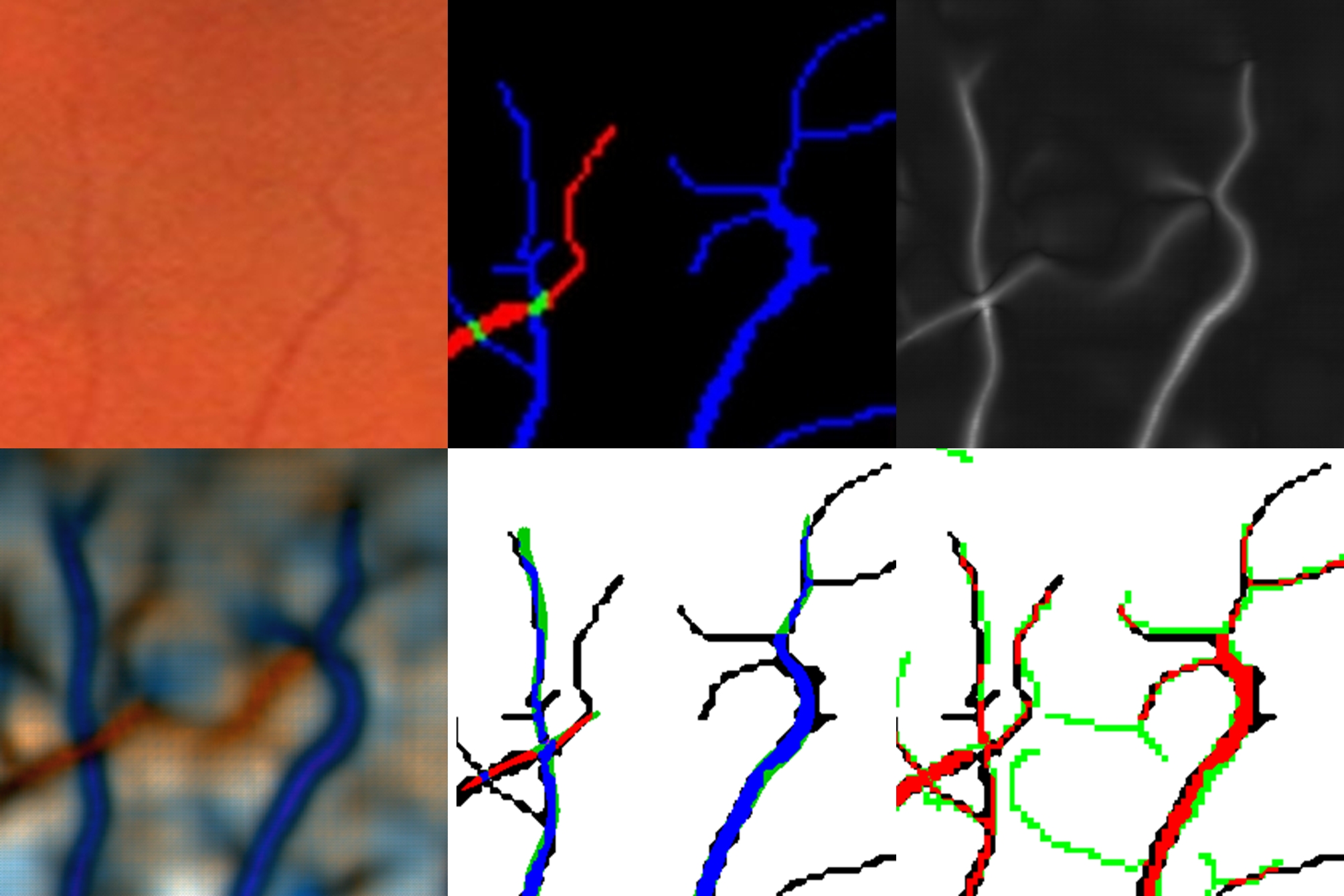}
\caption{Comparison of errors with a human expert. Top row from left to right: 1) original image, 2) ground truth, 3) Vessel probability map. Bottom row from left to right: 1) CNN output probability map (blue: veins; red: arteries; gray: background), 2) corresponding error map  (blue: true veins; red: true arteries; black: false negatives; green: false positives; white: true background), 3) error map of the second expert (red: true vessels; black: false negatives; green: false positives; white: true background). }
\label{fig_refst3}
\end{figure}

These errors are very similar to what human observers (in this case, the second expert) would make compared to the gold standard (see  Fig. \ref{fig_refst1}, Fig. \ref{fig_refst2} and Fig. \ref{fig_refst3}).

\begin{figure}[!t]
\centering
\includegraphics[width=3.4in]{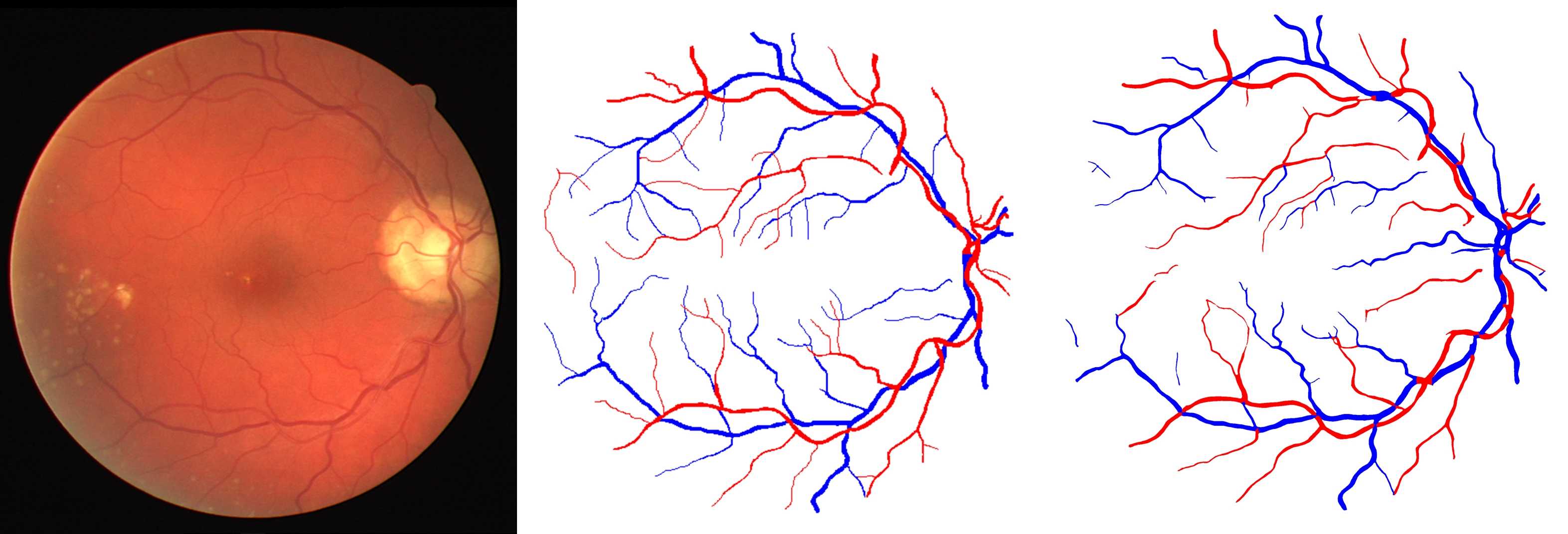}
\caption{Example of errors in presence of localized white spots. From left to right: 1) original image, 2) ground truth, 3) Vessel hard classification map }
\label{fig_newcase1}
\end{figure}

\begin{figure}[!t]
\centering
\includegraphics[width=3.4in]{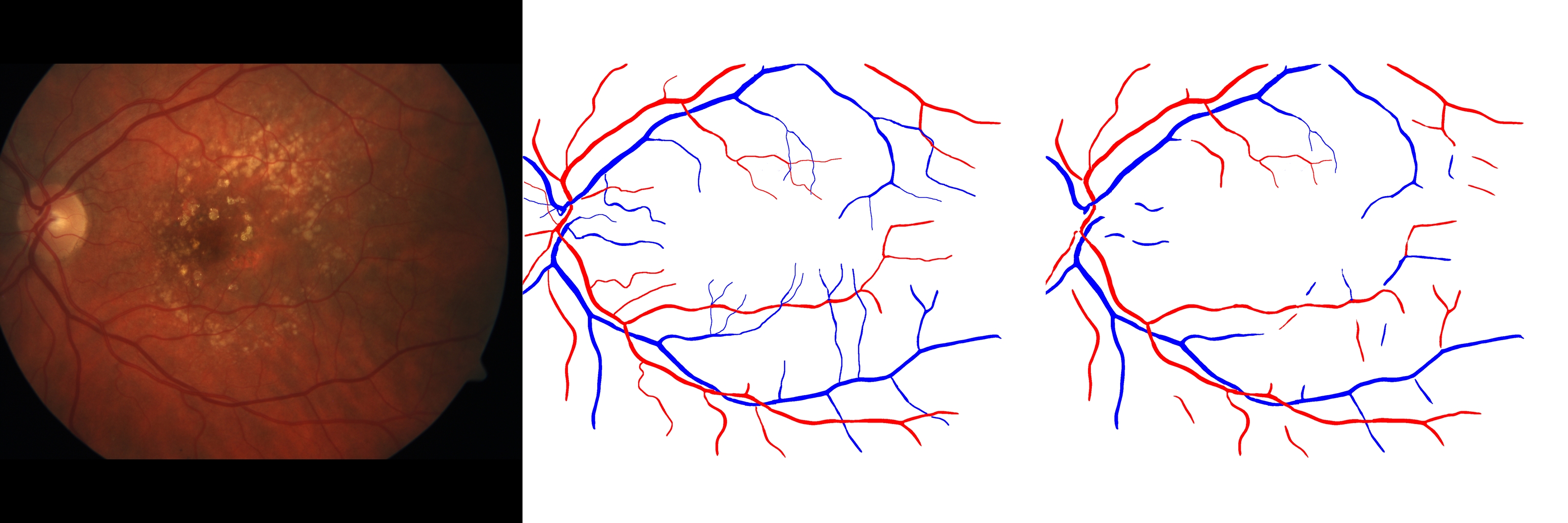}
\caption{Example of errors in presence of spread drusens due to moderate age-related macular degeneration (AMD).  From left to right: 1) original image, 2) ground truth, 3) Vessel hard classification map }
\label{fig_newcase2}
\end{figure}

\begin{figure}[!t]
\centering
\includegraphics[width=3.4in]{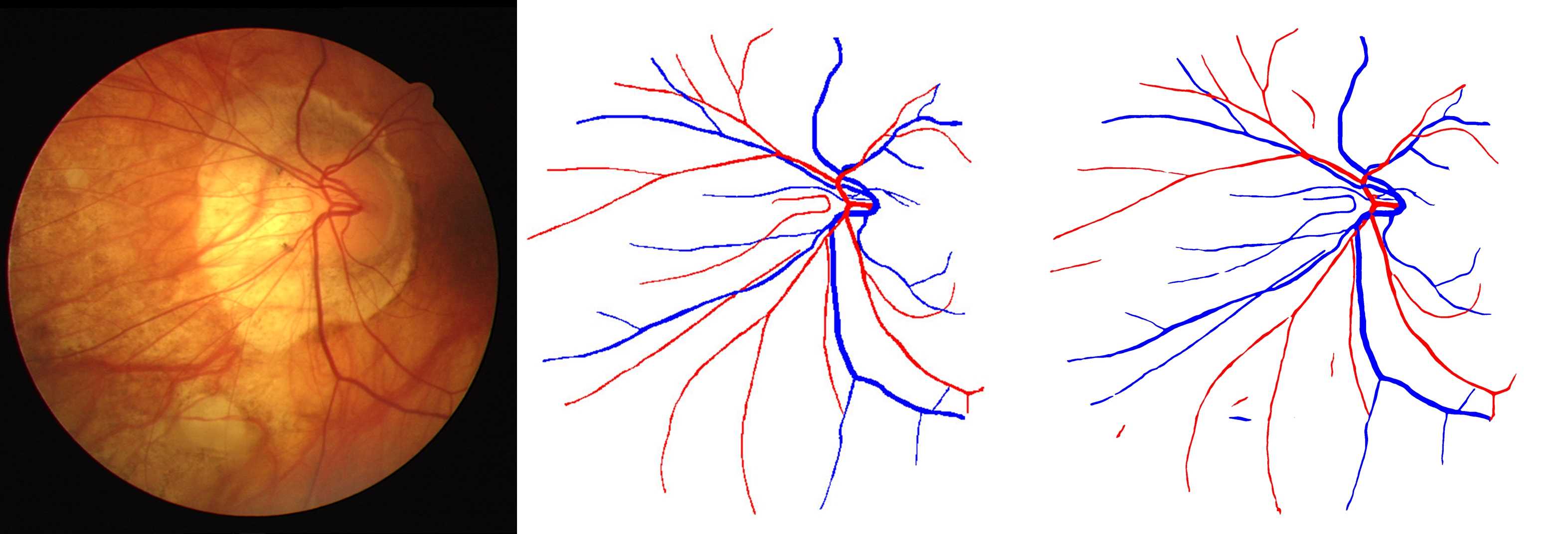}
\caption{Example of errors in presence of large white lesions.  From left to right: 1) original image, 2) ground truth, 3) Vessel hard classification map}
\label{fig_newcase3}
\end{figure}

Pathological condition can also complicate artery/vein classification especially when vessels are locally occluded by a lesion. We can observe the performance of our method in some cases where large or spread white lesions occurs due to diabetic retinopathy or age-related macular degeneration. In Fig.\ref{fig_newcase1}, the vessels that are on the borders of the lesions are not detected and a part of a detected vessel is misclassified. When the vessels are not occluded such as in Fig.\ref{fig_newcase2} and Fig. \ref{fig_newcase3}, the vessels are well classified while located in the lesion areas. Overall, the performance on large vessels is not affected.

In case of large/spread bright/dark lesions, if the lesion occludes the vessels, as in the first image in Fig.16, the occluded vessels will be absent or wrongly classified. If it is not occluded like in the second image and the third image (Fig.17 and Fig.18), the accuracy is not so much affected, one can see the small vessel are still difficult to detect but where there are white lesions, the model can still classify accurately the vessels.

\subsection{Comparison of segmentation results with the state of the art}

On the DRIVE dataset, the gold standard for vessel segmentation is based on the first expert who manually labeled the images. The second expert is considered as the human observer. However, we think that reporting the results of our system in comparison to both experts demontrates the good generalization capacity of our method. Indeed, as mentioned above, the second expert is completely independent from the training annotation, which is not the case for the first expert \cite{staal}.

We extracted the results for all the methods reported here from their respective papers (see Table \ref{table7}).

\begin{table}[!t]\caption{Comparison with the state of the art for vessel segmentation with the first expert considered as gold standard\label{table7}}
\centering
\resizebox{0.46\textwidth}{!}{%
\begin{tabularx}{0.59\textwidth}{c *{7}{s}}
\toprule
Method & AUC & Sens. & Spec. & Acc. & Time & System \\
\midrule
Niemeijier et al.\cite{niemeijervess} & 0.930 & 68.9\% & 96.9\% & 94.2\% & - \\ 
Martinez-Perez et al.\cite{martina} & - & 71.7\% & 96.6\% & 93.4\% & - & - \\
Roychowdhury et al.\cite{roychowdhury} & 0.967 & 73.9\% & 96.7\% & \textit{94.9\%} & 3.9s & 2.6Ghz \\
Miri et al.\cite{miri} & - & 71.5\% & 97.6\% & 94.3\% & 50s & 3Ghz\\
Staal et al.\cite{staal} & 0.952 & 67.8\%& 98.3\% & 94.4\% & 15m  & 1Ghz\\
Budai et al.\cite{budai} & - & 64.4\% & 98.7\% & \textit{95.7\%} & 5s & 2.3Ghz\\
Marin et al.\cite{marin} &  0.959 & 70.7\% & 98.0\% & 94.5\% & 90s & 2.1Ghz\\
Mendonca et al.\cite{mendonca} & - & 73.4\% & 97.6\% & 94.5\% & 150s & 3Ghz\\
Ricci et al.\cite{ricci} & 0.963 &  77.5\% & 97.2\% & \textit{95.9\%} & - & - \\ 
Human observer\cite{staal} & - &  77.6\% & 97.2\% &94.7\%& - & - \\ 
DeepVessel\cite{Fu2016} & 0.942 & 76.0\% & 97.4\% & 94.7\% & 1.3s & K40\\
Proposed & 0.964 & 74.9\% & 97.7\% & 94.8\% & 0.5s & 2.4Ghz\\
Fraz et al.\cite{fraz2} & 0.975 & 74.1\% & 98.1\% & 94.8\% & 100s & 2.3Ghz\\
Liskowski et al.\cite{liskowski} & 0.979 & 78.1\%& 98.1\% & 95.3\% & 92 s & Titan\\ 
\bottomrule
\end{tabularx}}
\end{table}

\begin{figure}[!t]
\centering
\includegraphics[width=2.8in]{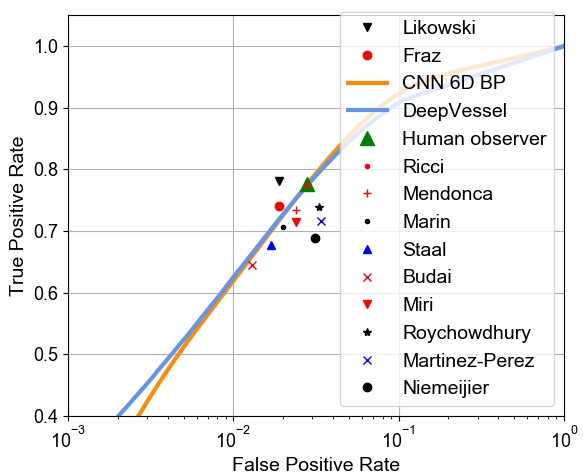}
\caption{Comparison with the state of the art for vessel segmentation (first expert).  }
\label{fig_rocstate}
\end{figure}

The AUC measure can be misleading since the performance for low specificity (high false positive rate) is not relevant. It is better to compare the different operating points obtained with all the compared methods againt the ROC curves for our method and DeepVessel (see Fig. \ref{fig_rocstate}). From this figure, we can see that the proposed method is similar to DeepVessel but below the Liskowski \cite{liskowski} and Fraz \cite{fraz2} methods. Both methods segment the vessels with binary pixel classification either with a features-based method \cite{fraz2} or a CNN-based method \cite{liskowski} which limits their speed and does not scale up well with high-resolution images. Their speed limitation is not easily tractable through optimization. The encoder-decoder strategy is much more efficient with a speed 200 times faster on a CPU.

\begin{figure}[!t]
\centering
\includegraphics[width=2.7in]{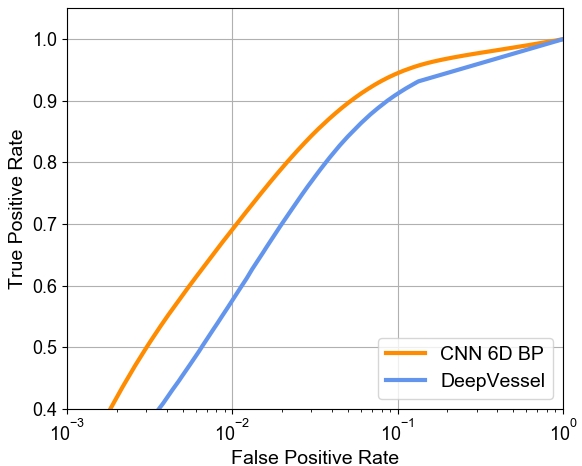}
\caption{Comparison with the state of the art for vessel segmentation (second expert).  }
\label{fig_resdeep}
\end{figure}

Most of the authors did not published their resulting segmentation images; for the DeepVessel method, however, the authors shared their output probability images, which allowed us to recompute their AUC and accuracy considering the second expert as the gold standard.
Table \ref{table8} and Fig. \ref{fig_resdeep} present results compared to the recent DeepVessel method using the second independent expert as gold standard.

\begin{table}[!t]\caption{Comparison with the state of the art for vessel segmentation with the second expert considered as gold standard\label{table8}}
\centering
\resizebox{0.46\textwidth}{!}{%
\begin{tabularx}{0.59\textwidth}{c *{7}{s}}
\toprule
Method & AUC & Sens. & Spec. & Acc. & Time &System\\
\midrule
DeepVessel\cite{Fu2016} & 0.947 & 72.7\% & 97.7\% & 94.6\% & 1.3s & K40\\
Proposed & 0.972 & 78.4\% & 98.1\% & 95.7\% & 0.5s & 2.4Ghz\\
\bottomrule
\end{tabularx}}
\end{table}

These results reveal that our system does not overfit to the training data; on the contrary, it generalizes well (better than DeepVessel, in fact), considering that the results in Table \ref{table8} and Fig. \ref{fig_resdeep} use a gold standard from a different expert than the one who labeled the training dataset. With the DRIVE database, we believe that system performance should always be evaluated against both experts' labelings of the test images, not just against the first, as this can reveal if a supervised learning method overfits to the way a particular expert has annotated the images.

These performances illustrate that our method achieves accuracy in segmenting vessels similar to the state of the art (within the inter-expert variability), with a shorter inference time. Our proposed CNN model is quite fast compared to the recent best method as instead of doing pixel classification, it uses a scalable encoding-decoding CNN model. The model can take the whole image directly as input, thereby avoiding the use of overlap strategies as with DeepVessel. Compared to the original U-Net model, patches fed to our network can be of any size, which is a significant advantage when dealing with images of different resolutions.

\subsection{Comparison of A/V classification results with the state of the art}

The classification performance for A/V classification is reported in Table \ref{table9}. The accuracy, sensitivity and specificity are computed at the pixel level. The positive detections are defined as the arteries pixels while the negative are veins. Table \ref{table9} also reports the performance of three recent state-of-the-art methods on CT-DRIVE, and of another CNN system we recently proposed \cite{fantin} which uses a pixel classification (PC) approach similar to \cite{ukbiobank, pulmonart}. Classifying arteries and veins is more accurate with the encoder-decoder model strategy and it is also significantly faster.
In our present method, the CNN step by itself gives good results and adding the LSP step significantly improves them.

\begin{table}[!t]\caption{Comparison with the state of the art for A/V classification on CT-DRIVE\label{table9}}
\centering
\resizebox{0.5\textwidth}{!}{%
\begin{tabularx}{0.55\textwidth}{Y d d b }
\toprule
 Method & Sensitivity & Specificity & Time\\
\midrule
CNN 6D BP + LSP  &  93.7\% $\pm$ 4\% &  92.9\% $\pm$ 5\% & 0.5s  \\
Estrada et al.\cite{estrada}* & 91.7\% $\pm$ 7\%& 91.7\% $\pm$ 7\% & 131.2s  \\
Dashbozorg et al.\cite{dasht} & 90.0\% & 84.0\% & -  \\
CNN 6D BP & 88.1\% $\pm$  5\% & 85.7\% $\pm$ 7\% & 0.4s \\ 
CNN PC \cite{fantin} & 86.0\% $\pm$  4\% & 83.8\% $\pm$ 9\% & 125s \\ 
Niemeijer et al.\cite{niemeijer}& 80.0\% & 80.0\% & - \\
\bottomrule
\end{tabularx}}
\end{table}

\subsection{Application to AVR measurement and novel global AVR measure}

Distinguishing arteries from veins allows to process statistics on the ratio of their diameters. Changes in the AVR are related to signs of hypertension, DR and other cardiovascular pathologies and thus we can make use of the joint segmentation and classification method to analyze changes in vessel diameters.
To evaluate the capability of AVR measures to analyze those changes, we use the MESSIDOR dataset, which contains 1200 fundus images. Three ophtalmologists provided the retinopathy grade for each image according to the number of microaneurysms, the number of hemorrhages and the presence of neovascularization. Grade 0 means that there is no microaneurysm nor hemorrhage; grade 1 means that there are fewer than 5 microaneurysms with no hemorrhages (similar to early DR in the standard classification); grade 2 means that there are no more than 15 microaneurysms and fewer than 5 hemorrhages (similar to moderate DR); finally, grade 3 denotes the presence of more than 15 microaneurysms or more than 5 hemorrhages, or the presence of neovascularization (proliferative DR).

The standard AVR measure is calculated with the algorithm proposed in \cite{niemeijer} and the revised ratio formulas from \cite{knudtson}. The conventional procedure is to calculate the ratio of Central Retina Artery Equivalent (CRAE) over Central Retina Vein Equivalent (CVAE) in the region between 0.5 disc diameters (DD) and 2 DDs away from the optic disc. The formulas from \cite{knudtson} are used to calculate the CRAE and CRVE. Only the six widest arteries and veins are retained. We will call this conventional measurement the Local AVR, as it uses only local information around the optic disc.
With automatic A/V classification applied to the whole fundus image, it is now possible to calculate an AVR over the whole field of view. We define a novel measurement, called Global AVR, as simply the ratio of the average of the artery diameters over the average of the vein diameters. To avoid any bias due to the fundus FOV, the global AVR should always be measured over the same portion of the retina. 
In the MESSIDOR dataset, this was not an issue as all the images are macula-centered and all have similar FOVs. The vessels map diameters were obtained using the method described in \cite{girardomia} and the optic discs were detected with the method described in \cite{girard2}.

\begin{figure}[!t]
\centering
\includegraphics[width=2.85in]{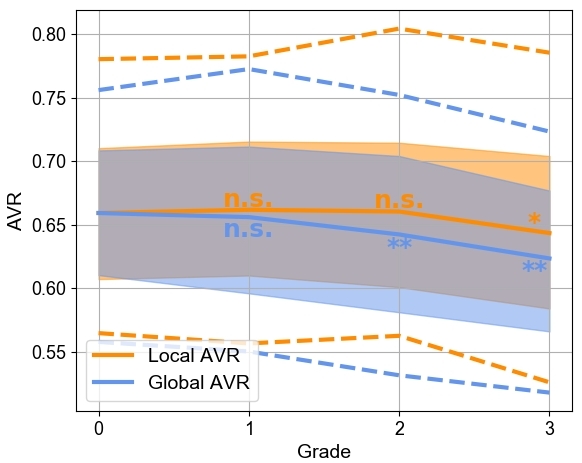}
\caption{Comparison between Local AVR and Global AVR for the different DR grades: n.s. means no significant difference, * means a significant p-value of  p$<$10$^{-4}$, ** means a significant p-value of  p$<$10$^{-5}$. Full lines are the means; shaded areas represent the data within 1 standard deviation, i.e. the [16\%-84\%] percentiles; dashed lines are the data within 2 standard deviations, i.e. the  [2.5\%-97.5\%] percentiles.}
\label{fig_avr1}
\end{figure}

We report in Table \ref{table10} the statistics for the Local and Global AVR for different DR grades as well as the number of cases for each severity level. These statistics are also represented graphically in Fig. \ref{fig_avr1}. 

\begin{table}[!t]\caption{AVR measures for different grades of DR on MESSIDOR\label{table10}}
\centering
\resizebox{0.48\textwidth}{!}{%
\begin{tabularx}{0.55\textwidth}{c *{5}{s}}
\toprule
 Method & Grade 0 & Grade 1 & Grade 2 & Grade 3\\
\midrule
\textit{Number} & 546 & 153 & 247 & 254 \\
Local AVR &  0.65$\pm$0.06  &   0.66$\pm$0.06$^{n}$  &   0.66$\pm$0.07$^{n}$  &  0.64$\pm$0.07$^{*}$  \\
Global AVR & 0.66$\pm$0.05  &   0.65$\pm$0.05$^{n}$  &   0.64$\pm$0.06$^{**}$ &  0.62$\pm$0.06$^{**}$ \\
\bottomrule
\multicolumn{5}{c}{Mann-Whitney significance test $^{n}$: p$>$0.1, $^{*}$: p$<$10$^{-4}$, $^{**}$: p$<$10$^{-5}$}
\end{tabularx}}
\end{table}

These results are coherent with those of the Wisconsin study \cite{avrdr}, which only shows statistically significant correlations between increased vein diameters (i.e. smaller local AVR) and proliferative DR (grade 3). On the other hand, using the Global AVR measure, we obtain significant differences not only for grade 3 but also for grade 2 which corresponds to moderate DR. We can also observe the mean curve for Global AVR (blue) begins to decrease already at grade 1 (early DR), while the Local AVR curve (orange) is increasing at this point.
The Local AVR is the measure used in the majority of clinical studies. In those studies, the AVR decreases with blood pressure increase,
but shows no significant change associated with DR at stages other than proliferative. The results here show clearly that a global measure should be considered in these studies as the difference in AVR will be more significant for proliferative DR and will also be significant for moderate DR. The reason for this could be that the earliest vessel changes, especially veinule dilation, appear on the smaller vessels further away from the OD, not in the area around the OD considered in the calculation of the Local AVR. 
This suggests that the whole vasculature visible in the fundus images contains important information for DR assessment, but only part of it is exploited by the standard AVR measure. Furthermore, detection of the optic disc is not a prerequisite for calculating our novel AVR measure.
The Global AVR is also more robust when the visibility of vessels
around the optic disc is altered either by lower image quality or pathologies. Around 10\% of grade 0 fundus images have AVR measures below 0.6. Among these cases, 60\% present factors that could explain the low AVR value (signs of old age, hypertension or cholesterol).

It is important to mention that factors that induce changes in the AVR can occur simultaneously and have opposite effects on the vessels diameters. For example, hypertension and obesity are associated with arteriolar narrowing and often occur simultaneously with diabetic retinopathy, itself associated with arteriolar widening \cite{avrdr}. Therefore, we must be cautious about our experimental results as the MESSIDOR dataset does not provide any normative data about age, cholesterol level, smoking or hypertension. Nonetheless, the present work shows that in future clinical studies, the Global AVR measure should be considered as a more reliable alternative to the standard local one. In addition, the fully automatic nature of our method can be an important advantage for carrying out future studies.

\subsection{Discussion and future work}
The experimental results show that our method outperforms the leading previous works in A/V classification on a public dataset (DRIVE). The vessel segmentation validation shows similar or better results than most of the state of the art methods. Our system is the fastest among all the methods compared, with all steps running on CPU and thus not requiring high-end and expensive GPUs to achieve short inference times. This is critical for the potential usefulness and impact of the method in a screening application.
The likelihood score propagation improves the performance on small vessels (less than 5 pixels wide), which is important because arteriolar narrowing and increased vessel tortuosity, that are related to cardiovascular pathologies, appears early in the smallest retinal vessels.
The performance on larger vessels exceeds 94\%; this means that our method can be used to obtain reliable standard (local) AVR mesurements fully automatically.

There is still room for improvement, especially as pertains to the CNN classification stage. Having more labeled data would obviously improve the classification performance. Indeed, this would allow our model to learn more global structural information by increasing the size of the training patches and the depth of the network. This would also allow the model to be more robust to severe pathologies that can complicate vessel classification.
In that respect, the proposed work speeds up manual A/V labeling as it provides an accurate first labeling that does not require a lot of manual corrections. In line with the need for labeled data, the labeled MESSIDOR training data, the label annotation tool implemented in this work, and the CNN model will be made available to the research community. 
The proposed work could also be used to automatically annotate new images and add then to the training dataset using an adversarial network. 
Future work will focus on finding new biomarkers and clinical measurements utilizing the artery/vein segmentation to track vessels changes and detect early signs of pathology.

\section{Conclusion}\label{sec:conclusion}

The novel and fast semantic artery/vein segmentation proposed in this paper combines a deep learning technique with a graph propagation method to jointly segment and classify vessels into arteries and veins. Our results confirm the efficiency of our novel training strategies (6D images, PCA augmentation and training patch selection). The CNN model used has the capacity to learn a wide variety of background characteristics and to distinguish retinal vessels from other tubular structures (choroidal vessels, optic disc border, nerve fibers), hence reducing false positive detections. Our use of global structure information is quite effective. Our technique for vessel labeling propagation mimics the natural blood flow in the vascular network and avoids learning a complex model with many topological rules. Moreover, we have shown that our method can be used to calculate a new arterio-venous ratio (AVR). The proposed Global AVR measure is better able to track vessel changes induced by diabetic retinopathy, which is very promising for the prospect of finding specific vascular biomarkers in more distal areas (further away from the optic disc). The potential impact of such a method is significant as it is fully automatic and therefore it could be used to screen patients for vascular changes that would need further attention.

\section*{Conflict of interest statement}

The authors declare that there are no conflicts of interest related to this article.

\section*{Acknowledgment}

This work was supported in part by the Natural Sciences and Engineering Research Council of Canada (NSERC). The authors would like to thank the research groups that made the DRIVE and MESSIDOR image database available to the community. The authors would like to thank Philippe Debann\'e for revising this manuscript, as well as Gabriel Lepetit-Aimon and Clement Playout for their technical input.


\bibliographystyle{elsarticle-num}

\bibliography{biblio_endnote}

\end{document}